\documentclass[sigconf]{acmart}

\usepackage{caption}

\usepackage{makecell}
\usepackage{booktabs,multirow,rotating}

\usepackage{colortbl}

\usepackage{ragged2e}

\usepackage{float}
\usepackage{subcaption}
\usepackage{hyperref}

\AtBeginDocument{%
  \providecommand\BibTeX{{%
    \normalfont B\kern-0.5em{\scshape i\kern-0.25em b}\kern-0.8em\TeX}}}

\copyrightyear{2022}
\acmYear{2022}
\setcopyright{acmcopyright}
\acmConference[MM '22] {Proceedings of the 30th ACM International Conference on Multimedia }{October 10--14, 2022}{Lisboa, Portugal.}
\acmBooktitle{Proceedings of the 30th ACM International Conference on Multimedia (MM '22), October 10--14, 2022, Lisboa, Portugal}
\acmPrice{15.00}
\acmISBN{978-1-4503-9203-7/22/10}
\acmDOI{10.1145/3503161.3547930}

\begin{document}

\title{Synthetic Data Supervised Salient Object Detection}


%


\author{Zhenyu Wu}
\affiliation{%
  \institution{State Key Laboratory of Virtual Reality Technology and Systems, Beihang University}
  \city{Beijing}
  \country{China}
}

\author{Lin Wang}
\affiliation{%
  \institution{School of Transportation Science and Engineering, Beihang University}
  \city{Beijing}
  \country{China}
}

\author{Wei Wang}
\affiliation{%
  \institution{School of Computer Science and Technology, Harbin Institute of Technology}
  \city{Shenzhen}
  \country{China}
}

\author{Tengfei Shi}
\affiliation{%
  \institution{State Key Laboratory of Virtual Reality Technology and Systems, Beihang University}
  \city{Beijing}
  \country{China}
}

\author{Chenglizhao Chen}
\authornote{Corresponding Author: Chenglizhao Chen, cclz123@163.com}
\affiliation{%
  \institution{College of Computer Science and Technology, China University of Petroleum (East China)}
  \city{Qingdao}
  \country{China}
}

\author{Aimin Hao}
\affiliation{%
  \institution{State Key Laboratory of Virtual Reality Technology and Systems, Beihang University, Beijing}
  \institution{Peng Cheng Laboratory, Shenzhen}
  \country{China}
}

\author{Shuo Li}
\affiliation{%
  \institution{Department of Medical Imaging, Western University}
  \city{London}
  \country{Canada}
}

\renewcommand{\shortauthors}{Zhenyu Wu et al.}


\begin{abstract}
Although deep salient object detection (SOD) has achieved remarkable progress, deep SOD models are extremely data-hungry, requiring large-scale pixel-wise annotations to deliver such promising results. In this paper, we propose a novel yet effective method for SOD, coined SODGAN, which can generate infinite high-quality image-mask pairs requiring only a few labeled data, and these synthesized pairs can replace the human-labeled DUTS-TR to train any off-the-shelf SOD model. Its contribution is three-fold. \textbf{1)} Our proposed diffusion embedding network can address the manifold mismatch and is tractable for the latent code generation, better matching with the ImageNet latent space. \textbf{2)} For the first time, our proposed few-shot saliency mask generator can synthesize infinite accurate image synchronized saliency masks with a few labeled data. \textbf{3)} Our proposed quality-aware discriminator can select highquality synthesized image-mask pairs from noisy synthetic data pool, improving the quality of synthetic data. 
For the first time, our SODGAN tackles SOD with synthetic data directly generated from the generative model, which opens up a new research paradigm for SOD.
Extensive experimental results show that the saliency model trained on synthetic data can achieve $98.4\%$ F-measure of the saliency model trained on the DUTS-TR. Moreover, our approach achieves a new SOTA performance in semi/weakly-supervised methods, and even outperforms several fully-supervised SOTA methods. Code is available at \href{https://github.com/wuzhenyubuaa/SODGAN}{https://github.com/wuzhenyubuaa/SODGAN}
\end{abstract}

\ccsdesc[500]{Computer methodologies~Artificial intelligence}
\ccsdesc[500]{Computer methodologies~Machine learning}

\keywords{Salient object detection, Synthetic data, Semi-supervised learning}


\maketitle

\section{Introduction}
\begin{figure}[t]
    \centering
    \begin{subfigure}{.11\textwidth}
      \centering
      \includegraphics[width=\textwidth]{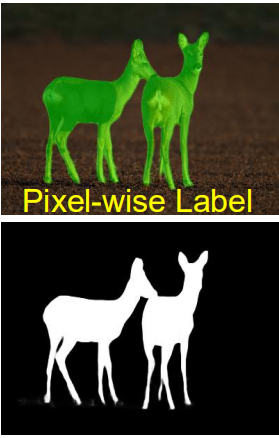}
      \caption{ PFSN~\cite{ma2021PFSNet}}
    \end{subfigure}\hspace*{-0.150em}
    \begin{subfigure}{.11\textwidth}
      \centering
      \includegraphics[width=\textwidth]{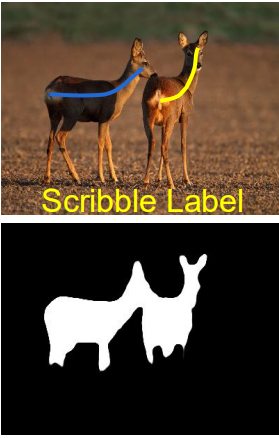}
      \caption{ SCWS \cite{yu2021structure}}
      \label{fig:scws}
    \end{subfigure}\hspace*{-0.15em}
     \begin{subfigure}{.11\textwidth}
      \centering
      \includegraphics[width=\textwidth]{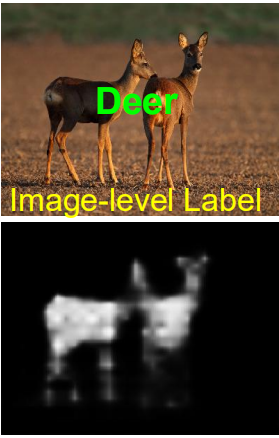}
      \caption{ MWS \cite{zeng2019multi}}
      \label{fig:mws}
    \end{subfigure}\hspace*{-0.15em}
     \begin{subfigure}{.11\textwidth}
      \centering
      \includegraphics[width=\textwidth]{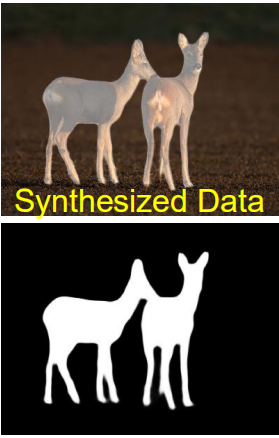}
      \caption{ \textbf{Ours}}
      \label{fig:ours}
    \end{subfigure}\vspace{-1em}
    \caption{The saliency model trained on synthetic data outperforms SOTA weakly-supervised methods, and is even competitive with fully-supervised models.} \vspace{-1.2em}

    \label{fig:our_vs_others}
\end{figure}

Salient object detection (SOD) aims to segment interesting objects that attract human attention in an image. As a fundamental tool, it can be leveraged to various applications including scene understanding \cite{Zhou_2021_ICCV}, semantic segmentation \cite{Zhou_2020_CVPR} and image editing \cite{Jiang_2021_CVPR,cheng2010repfinder}.
Recently, SOD has achieved significant progress \cite{Fang_2021_ICCV,Liu_2021_ICCV,CC_2021_CVPR,Tang_2021_ICCV,zhao2021complementary,zhang2021auto,wu2022recursive} due to the development of deep model. However, deep networks are extremely data-hungry, typically requiring pixel-level humanannotated datasets to achieve high performance (see Fig. \ref{fig:our_vs_others}.a). Labeling large-scale datasets with pixel-level annotations for SOD is very time-consuming, e.g., generally more than five people were asked to annotate the same image to guarantee the label consistency and another ten viewers were asked to cross-check the quality of annotations in the SOC dataset \cite{fan2018salient}.

To alleviate the dependency on pixel-wise annotation, many weakly-supervised SOD methods~\cite{zeng2019multi,li2018weakly,wang2017learning} have been devised. Typically, image-level labels (see Fig. \ref{fig:our_vs_others}.c) are utilized in \cite{li2018weakly,wang2017learning} for saliency localization, and then iteratively finetune their models with predicted saliency maps. Additionally, scribble annotations (see Fig. \ref{fig:our_vs_others}.b) has been proposed recently in \cite{zhang2020weakly} to reduce the uncertainty of image-level labels.
Although these methods are free of pixel-level annotations, they suffer from various disadvantages, including low prediction accuracy, complex training strategy, dedicated network architecture, and extra data information (e.g., edge) to obtain high-quality saliency maps.


In this paper, we propose a new paradigm SODGAN (see Fig. \ref{fig:our_vs_others}.d) for SOD, which can generate infinite high-quality image-mask pairs with a few labeled data to replace the human-labeled DUTS-TR \cite{wang2017learning} dataset. Concretely, our SODGAN has three stages: \textit{Stage 1}. Learning a few-shot saliency mask generator to synthesize image-synchronous mask, while utilizing the existing generative adversarial networks (BigGAN \cite{brock2018large}) to generate realistic images. \textit{Stage 2}. Selecting high-quality image-mask pairs from the synthetic data pool. \textit{Stage 3}. Training a saliency network on these filtered image-mask pairs. However, there are three main challenges with this approach: \textbf{1) Lacking pixel-wise labeled data} as the training dataset to learn a segmentor because BigGAN was trained on the ImageNet that was designed to classification tasks without the pixel-level label. \textbf{2) Discovering a meaningful direction} in GAN latent space to disentangle foreground saliency objects from backgrounds is nontrivial, which often requires domain knowledge and laborious engineering. \textbf{3) Low-quality image-mask pairs} exist in the synthesized datasets.

To tackle these three challenges, \textbf{first}, we present a diffusion embedding network (DEN) (see Sec. 3.2) to utilize the existing well-annotated dataset (i.e., DUTS-TR), which can infer the image’s latent code that match with the ImageNet latent code space; thus, the existing labeled DUTS-TR dataset can provide the pixel-wise label for ImageNet. \textbf{Second}, in contrast to the existing works \cite{shen2020interpreting, goetschalckx2019ganalyze, plumerault2019controlling} focusing on latent space, we propose a few-shot saliency mask generator to automatically discover meaningful directions in the GANs feature space (see Sec. 3.3), which can synthesize infinite high-quality image synchronized saliency masks with a few labeled data. \textbf{Third}, we propose a quality-aware discriminator (see Sec. 3.4) to select high-quality synthesized image-mask pairs from the noisy synthetic data pool, improving the quality of synthetic data.

Our SODGAN has several desirable properties. \textbf{a) Fewer labels.} Our approach eliminates large-scale pixel-level supervision requiring only a few labeled data, which reduces the annotation costs. \textbf{b) High performance.} We demonstrate that the saliency model trained on synthetic data directly generated from GANs achieves an average $98.4\%$ F-measure of the saliency model trained on the DUTS-TR dataset. Moreover, our SODGAN achieves new SOTA performance in semi/weakly-supervised methods, and even outperforms some fully supervised methods. \textbf{c) Generality.} The synthetic data can be used to train any off-the-shelf SOD model without the need of special architectures, showing strong generalization capabilities on the real test datasets. We summarize the key contributions as follows:\vspace{-1em}

\begin{itemize}
\item For the first time, our SODGAN tackles SOD with synthetic data directly generated from the generative model, which opens up a new research paradigm for semi-supervised SOD and significantly reduces the annotation costs.
\item Our proposed the DEN can address manifold mismatch and is tractable for the latent code generation, better matching with the ImageNet latent space.
\item Our lightweight few-shot saliency mask generator can synthesize infinite accurate image-synchronous saliency masks with a few labeled data.
\item Our proposed quality-aware discriminator can select highquality synthesized image-mask pairs from the noisy synthetic data pool, improving the quality of synthetic data.
\end{itemize}

\section{Related Work}
\noindent{\textbf{Semi/Weakly-supervised SOD Approaches.}} With recent advances in semi/weakly-supervised learning, a few existing works exploit the potential of training saliency detectors on image-level~\cite{zeng2019multi,li2018weakly,wang2017learning}, region-level~\cite{yu2021structure, zhang2020weakly,zhang2020learning}, and limited pixel-level~\cite{ zhang2021few, wu2020deeper,yan2019semi,zhou2018semi} labeled data to relax the dependency of manually annotated pixel-level saliency masks. For image-level supervision, these approaches \cite{zeng2019multi,li2018weakly,wang2017learning} follow the same technical route, \emph{i.e.}, producing initial saliency maps with image-level labels and then further refining it via iterative training. Recently, scribble annotation was proposed in~\cite{zhang2020weakly,yu2021structure}, but it requires large-scale scribble annotations (10553 images) and extra data information (\emph{e.g.}, edge) to recover integral object structure.
\textbf{Differences.}
Distinct from all these works, our approach provides a new paradigm for semi-supervised SOD. In particular, we introduce SODGAN, a generative model, which can generate infinite high-quality image-mask pairs requiring minimal manual intervention. These generated pairs can then be used for training any existing SOD approaches.

\noindent{\textbf{Latent Interpretability of GANs.}} The previous works have shown that the GANs latent spaces are endowed with human-interpretable semantic arithmetic. A line of recent works~\cite{shen2020interpreting, goetschalckx2019ganalyze, plumerault2019controlling,shen2021closed,cherepkov2021navigating,yang2021discovering} employ explicit human-provided supervision to identify interpretable directions in the latent space. For instance, \cite{goetschalckx2019ganalyze, shen2020interpreting} use the classifiers pretrained on the CelebA~\cite{liu2015deep} dataset to produce pseudo labels for the generated images and their latent codes.
Another active line of study on GANs~\cite{abdal2021labels4free,chen2019unsupervised, bielski2019emergence, melas2021finding,voynov2021object,zhang2021datasetgan,tritrong2021repurposing} targets the object segmentation task. \cite{abdal2021labels4free} and \cite{chen2019unsupervised} are based on the idea of decomposing the generative process in a layer-wise fashion.
Other works~\cite{bielski2019emergence, melas2021finding,voynov2021object} exploit the idea that the object’s location or appearance can be perturbed without affecting image realism.
\textbf{Differences.}
In contrast to existing works manipulating the latent space, our approach is able to discover interpretable directions in the GANs features space, which allows complete control over the diversity of object categories and can automatically find the expected directions.

\begin{figure*}[t]
\centering
\includegraphics[width=0.97\textwidth]{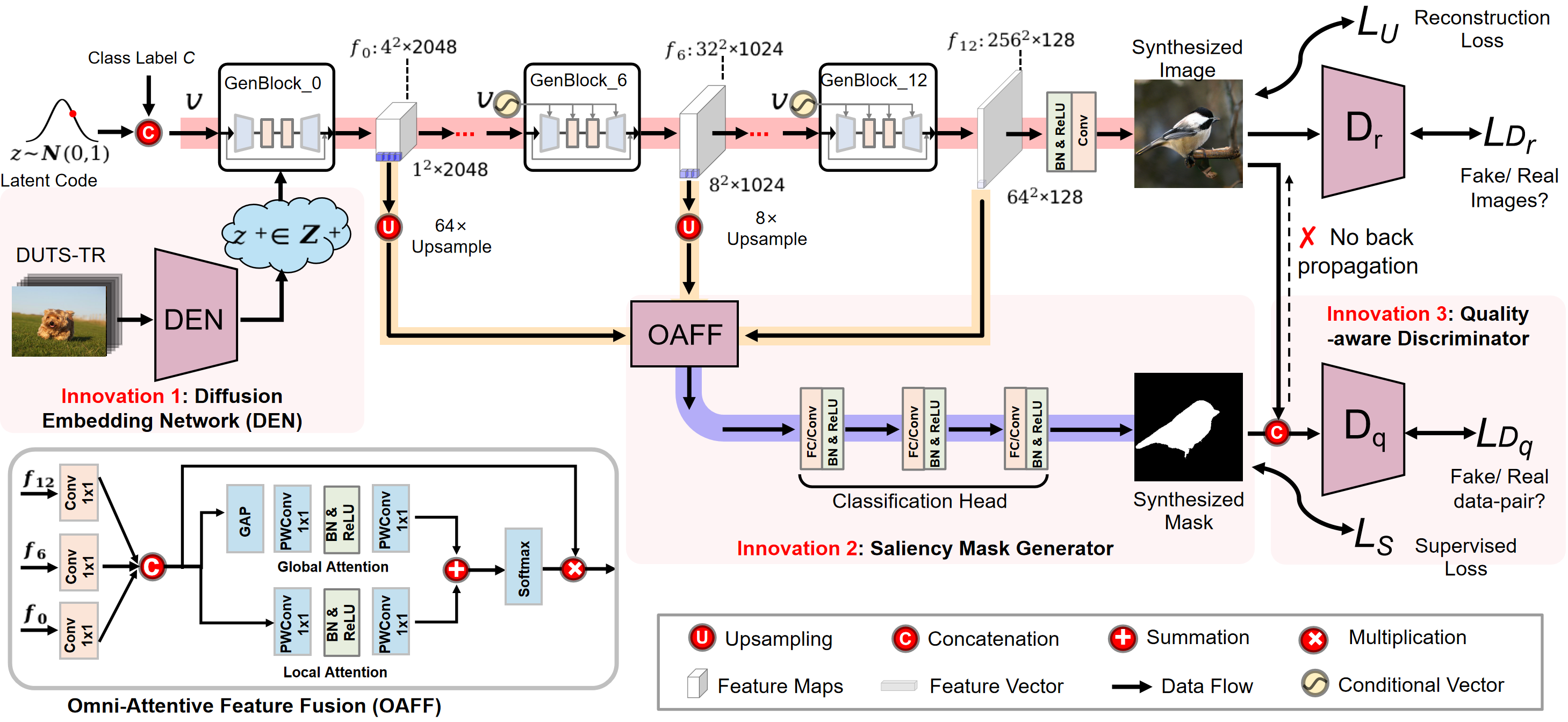}
\vspace{-0.6em}
\caption{Overview of the proposed SODGAN. Given a latent code $z \sim \mathcal{N}(0, 1)$ and a class label $c$, we collect hidden semantic feature $\{f_0, f_1, ..., f_{12}\}$ from $G_{image} (\cdot)$ to disentangle foregrounds from backgrounds. After that, we upsample these collected feature maps to $256 \times 256$ resolution and then concatenate these upsampled features together, constructing pixel-wise feature maps for all pixels of the synthesized image. Finally, these pixel-wise features are fed into the proposed saliency mask generator branch to produce image-synchronized saliency mask.}
\label{fig:pipeline}
\end{figure*}

\section{Method}
\label{method}

\subsection{Overview}

As shown in Fig. \ref{fig:pipeline}, our SODGAN is composed of the diffusion embedding network $DEN(\cdot)$, the mask synthesis network $G_{mask}(\cdot)$, the quality-aware discriminator $D_{q}(\cdot)$, the image synthesis network $G_{image}(\cdot)$, and the image reconstruction discriminator $D_{r}(\cdot)$. \textbf{(1)} The proposed $DEN(\cdot)$ aims to address the lacking of pixel-wise labels in ImageNet, which is designed for recognition tasks without segmentation groundtruth. Our $DEN(\cdot)$ can utilize existing labeled datasets DUTS-TR \cite{wang2017learning}, and gradually turn it into a unique latent space $Z+$ that matches with the ImageNet latent code space. \textbf{(2)} The proposed $G_{mask}(\cdot)$ is to discover meaningful directions in the GANs feature space, synthesizing image synchronized saliency mask. Our $G_{mask}(\cdot)$ is build on top of the $G_{image}(\cdot)$ architecture augmented with a few-shot saliency mask generation branch, \textbf{(3)} Our $D_{q}(\cdot)$ is designed to select high-quality synthesized image-mask pairs from noisy synthetic data pool. The $G_{image}(\cdot)$ can be any off-the-shelf GANs models, and the $D_{r}(\cdot)$ is the corresponding real/fake discriminator. Here, we demonstrate our approach using BigGAN \cite{brock2018large}, a class-conditional GANs trained on ImageNet \cite{deng2009imagenet}. In our SODGAN, the proposed $DEN(\cdot)$, $G_{mask}(\cdot)$ and $D_{q}(\cdot)$ are trainable while the other components remain fixed.

\subsection{Diffusion Embedding Network}

Our $DEN(\cdot)$ is to address the lacking of pixel-wise label in ImageNet, which is designed for recognition tasks without segmentation groundtruth, better matching with ImageNet latent code space. Previous work \cite{zhang2021datasetgan} addresses this issue by manually labeling a handful of sampled images, which is labor-consuming. An alternative idea is to utilize the existing labeled datasets (e.g., DUTS-TR) by using variational autoencoder (VAEs). However, the standard VAEs, with a Euclidean latent space, is structurally incapable of capturing topological properties of certain datasets, which is called manifold mismatch \cite{falorsi2018explorations}.

To address these challenges, we developed the diffusion embedding network $DEN(\cdot)$ to utilize the existing labeled datasets with pixel-wise annotation (e.g., DUTS-TR), which allows for an arbitrarily closed manifold as a latent space and captures the underlying geometrical structure. The proposed $DEN(\cdot)$ can gradually turn an image into a unique latent code $z^+$ that better matches with ImageNet latent code space. Concretely, our $DEN(\cdot)$ are latent variable models of the forms $p_{\theta}(x_0)=\int p_{\theta}(x_{0:T})dx_{1:T}$, where $x_1, ..., x_T$ are intermediate latent codes and $x_0 \sim q(x_0)$ is the initial image. The joint distribution $p_{\theta}(x_{0:T})$ is the embedding process, and it is defined as the Markov chain with learned Gaussian transitions $p(x_T) = \mathcal{N}(x_T;0,1)$:

\begin{equation}
\begin{aligned}
p_{\theta}\left(x_{0: T}\right) &=p\left(x_{T}\right) \prod_{t=1}^{T} p_{\theta}\left(x_{t-1} \mid x_{t}\right), \\
p_{\theta}\left(x_{t-1} \mid x_{t}\right) &=\mathcal{N}\left(x_{t-1} ; \mu_{\theta}\left(x_{t}, t\right), \sum_{\theta}\left(x_{t}, t\right)\right)
\end{aligned}
\end{equation}
The difference between our $DEN(\cdot)$ and VAEs is that the approximate posterior $q(x_{1:T}|x_0)$, which is called the diffusion process, is fixed to a Markov chain that progressively adds Gaussian noise to the image in line with variance schedule $\beta_1, ..., \beta_T$:

\begin{figure}[t]
\centering
\includegraphics[width=0.85\linewidth]{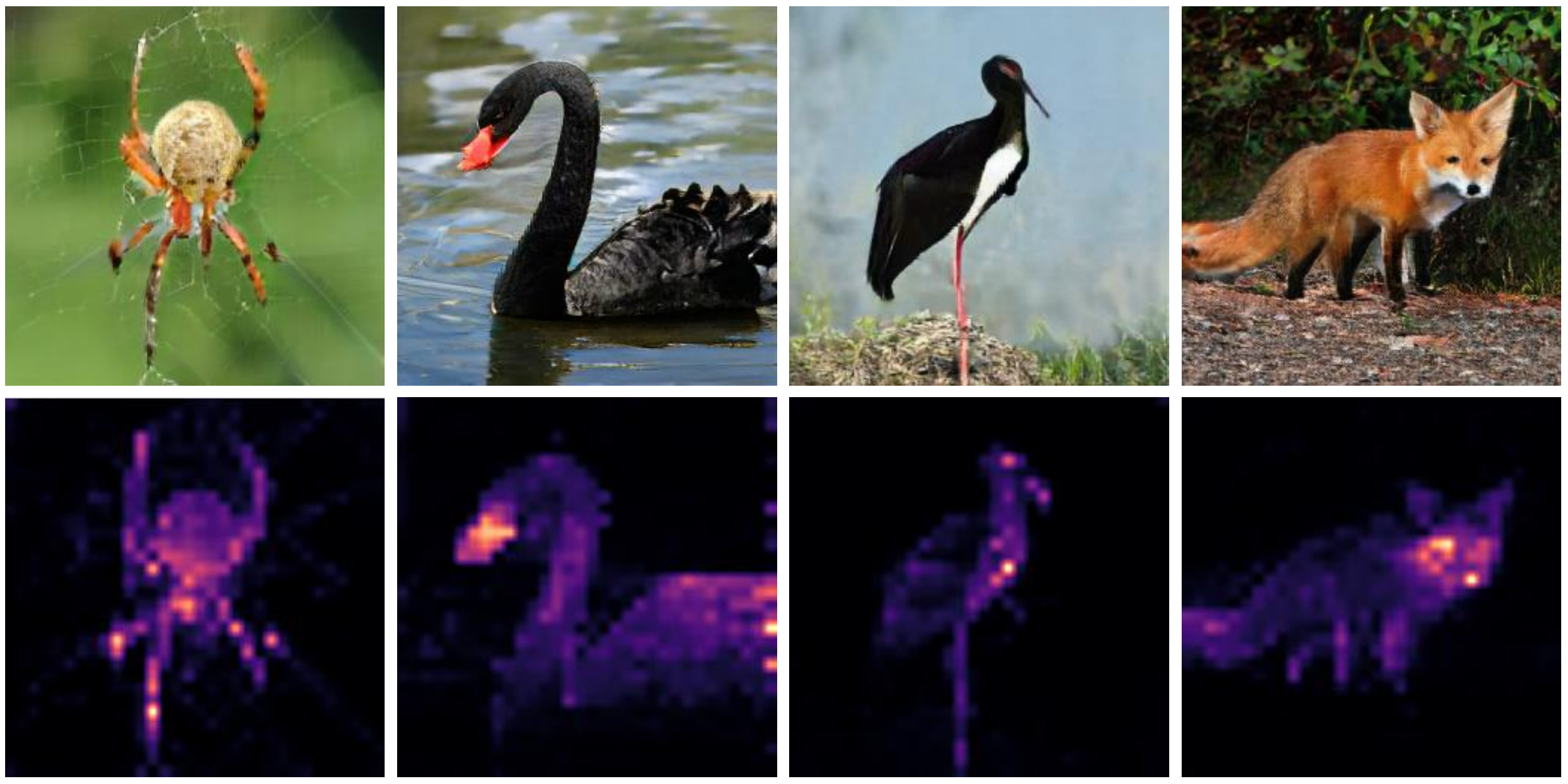}\vspace{-0.7em}
\caption{Visualizing the omni-attention maps of hidden features, which can locate the salient object masks coarsely.}
\label{fig:internal_feature}
\end{figure} \vspace{-1.7em}

\begin{equation}
\begin{aligned}
q\left(x_{1: T} \mid x_{0}\right) &=\prod_{t=1}^{T} q\left(x_{t} \mid x_{t-1}\right) \\
q\left(x_{t} \mid x_{t-1}\right) &=\mathcal{N}\left(x_{t} ; x_{t-1} \sqrt{1-\beta_{t}}, \beta_{t}\right)
\end{aligned}
\end{equation}
A desirable property of the diffusion process is that it admits sampling $x_t$ at a arbitrary timestep $t$ in closed form:

\begin{equation}
q\left(x_{t} \mid x_{0}\right)=\mathcal{N}\left(x_{t} ; x_{0} \sqrt{\hat{\alpha}}, 1-\hat{\alpha}\right)
\end{equation}
where $\alpha = 1 - \beta_t$ and $\hat{\alpha} = \prod_{s=1}^{t}\alpha_s$. The reconstruction loss is to optimize the variational bound on negative log likelihood:
\begin{equation}
\begin{aligned}
\mathcal{L}_{U}&= \mathbb{E}\left[-\log p_{\theta}\left(x_{0}\right)\right] \leq \mathbb{E}\Big[-\log \frac{p_{\theta}(x_{0: T})}{q(x_{1: T} \mid x_{0})}\Big] \\
&=\mathbb{E}\Big[\sum_{t>1} D_{K L}(q(x_{t-1} \mid x_{t}, x_{0}) \| p_{\theta}(x_{t-1 \mid x_{t}}))-\log _{\theta}(x_{0} \mid x_{1}).\\
&+D_{K L}(q(x_{T} \mid x_{0}) \| p(x_{T})\Big]
\end{aligned}
\end{equation}
where $D_{KL}(\cdot)$ is the KL divergence. The adversarial loss can be defined as:
\begin{equation}
\begin{aligned}
\mathcal{L}_{D_{r}} &=\mathbb{E}_{x \sim \mathcal{D}_{dut}}\left[\log \left(D_{r}(x)\right)\right] \\
&+\mathbb{E}_{x \sim \mathcal{D}_{dut}}\left[\log \left(1-D_{r}\left(G_{\text{image}}(E(x))\right)\right)\right]
\end{aligned}
\end{equation}
where $\mathcal{D}_{dut}$ is the DUTS-TR dataset. Note that our $DEN(\cdot)$ doesn’t need special architectures. Here we adopt the MobileNetV3 \cite{howard2019searching} architecture for the diffusion model. In this way, given an image, our $DEN(\cdot)$ can infer its latent code $z^+$ that matches with the ImageNet latent code space, and find its groundtruth $y$ in the DUTS-TR.

\noindent\textbf{Summarized Advantages: 1)} Gaussian noise has the effect of filling low density regions in the original data distribution; thus, our $DEN(\cdot)$ can obtain more training signal to improve latent distributions that faster converge to the true data distribution. \textbf{2)} Our $DEN(\cdot)$ is capable of capturing topological properties of certain datasets that better match with ImageNet latent code space.

\subsection{Few-shot Saliency Mask Generator}

Our $G_{mask}(\cdot)$ is a lightweight few-shot generator trained on a few labeled data, which can synthesize infinite image synchronized accurate saliency masks. The $G_{mask}(\cdot)$ consists of the omniattentive feature fusion module (OAFF), and the classification head, sharing the same feature extractor with the $G_{image}(\cdot)$. Let the $f_i \in \mathbb{R}^{W_i \times H_i \times C_i}$ denote the hidden representation of $G_{image}(\cdot)$.

\noindent \textbf{Omni-Attentive Feature Fusion.} Previous work \cite{yang2021discovering} has demonstrated that in GANs feature space, low-level features contain local information like texture and color while high-level features capture global information, such as the style and layout of objects. To fully take advantage of the multi-level features, we proposed a novel omni-attentive feature fusion module, as depicted in Fig. \ref{fig:pipeline} (bottomleft). To ensure the spatial alignment, we first upsample all feature maps $\{f_0, f_1, ..., f_{l}\}$ to the highest output resolution $256 \times 256$, and then concatenate them along the channel dimension to obtain an aggregated feature $f^*$:
\begin{equation}
f^{*}=\operatorname{Cat}\Big(\operatorname{Conv}_{1 \times 1}(U(f_{0})), \operatorname{Conv}_{1 \times 1}(U(f_{1})), ..., \operatorname{Conv}_{1 \times 1}(U(f_{l}))\Big)
\end{equation}
where $U(\cdot)$ denotes upsample operation, $\operatorname{Conv}_{1 \times 1}(\cdot)$ is $1 \times 1$ convolutional operation for reducing channel dimension, and $\operatorname{Cat}(\cdot)$ stands for concatenation. To better fusion the global and local contexts, we introduce the omni-attention ${OA}(\cdot)$ module, including local attention $LA(\cdot)$ and global attention $GA(\cdot)$:

\begin{equation}
\begin{aligned}
OA\left(f^{*}\right) &=LA\left(f^{*}\right)+GA\left(f^{*}\right) \\
LA\left(f^{*}\right) &=PWC\left(\operatorname{ReLU}\left(BN\left(PWC\left(f^{*}\right)\right)\right)\right) \\
GA\left(f^{*}\right) &=PWC\left(\operatorname{ReLU}\left(BN\left(PWC\left(GAP\left(f^{*}\right)\right)\right)\right)\right)
\end{aligned}
\end{equation}
where the $GAP(\cdot)$ is global average pooling, and $PWC(\cdot)$ is the $1 \times 1$ point-wise convolution for reducing the parameters. Fig. \ref{fig:internal_feature} shows the visualized omni-attention maps. The aggregated feature $f'$ can be obtained by multipling with the $OA(f^*)$:

\begin{equation}
f^{\prime}=f^{*} \otimes OA\left(f^{*}\right)
\end{equation}
where the $\otimes$ is the element-wise multiplication operator.

\noindent \textbf{Classification Head.} After that, the  $f' \in \mathbb{R}^{256 \times 256 \times C}$ are flattened into vectors $X \in \mathbb{R}^{65536 \times C}$, constructing pixel-wise feature vectors for all pixels on the synthesized image. Finally, we feed it into the proposed classification head network to produce pixel-wise saliency mask, where the detailed network structure discussed in Sec. 4.1.

To train $G_{mask}(\cdot)$, we need to collect a small training set $\mathcal{D}_m = \{(z^+_1, y_1),..., ((z^+_k, y_k)) \}$, where $y_i$ is selected from the DUTS-TR. Specifically, we use the state-of-the-art (SOTA) image classification model CoAtNet \cite{dai2021coatnet} to classify the DUTS-TR, which can be divided into 522 categories. We then randomly select a pair of $(z^+, y)$ for each class, forming a small training set with 522 images. We then train the proposed $G_{mask}(\cdot)$ by using activation features $z^+$ and the corresponding pixel-wise annotations. The training objective is
\begin{equation}
\mathcal{L}_{G_{\text {mask }}}=\mathcal{L}_{S}+\mathcal{L}_{D_{q}}
\end{equation}
$\mathcal{L}_{S}$ is the supervised loss on labeled images with a combination of cross entropy and dice loss, defined as:
\begin{equation}
\mathcal{L}_{S}=\frac{1}{H W} \sum_{i}^{H} \sum_{j}^{W} y_{i j} \log \left(\hat{y}_{i j}\right)+\left(1-\frac{2 \sum_{i}^{H} \sum_{j}^{W} y_{i j} \hat{y}_{i j}}{\sum_{i}^{H} \sum_{j}^{W}\left[y_{i j}+\hat{y}_{i j}\right]}\right)
\end{equation}
where the $H$ and $W$ are the height and width of the image respectively, and $\hat{y}_{ij}$ is the prediction probability at position $(i,j)$. The quality-aware discriminator loss $\mathcal{L}_{D_q}$ is: 

\begin{figure*}[t]
\centering
\includegraphics[width=0.95\textwidth]{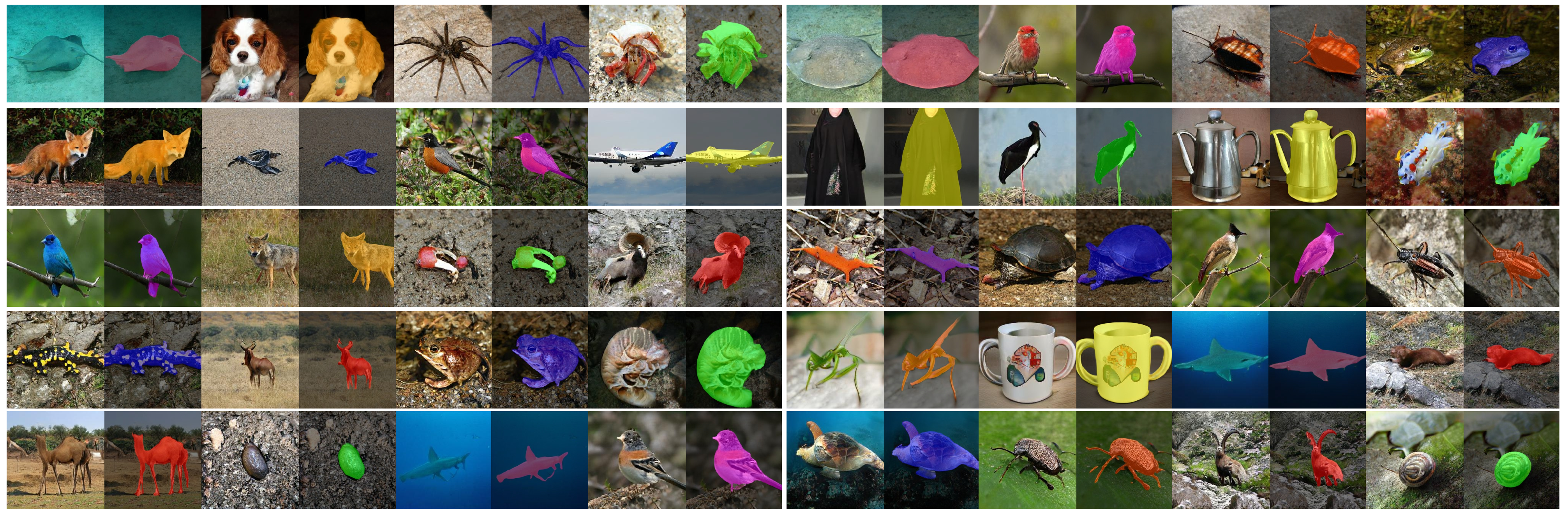}
\vspace{-1em}
\caption{With a few labeled data, our SODGAN can generate infinite realistic and diverse (1,000 categories) image-mask pair.}
\label{fig:gan_pic}
\end{figure*}

\begin{equation}
\begin{aligned}
\mathcal{L}_{D_{q}}&=\operatorname{argmin}_{G_{\operatorname{mask}}} \mathbb{E}_{(x_{{real }}, y_{{real }}) \sim \mathcal{D}_{m}}[\log (D_{q}(x_{{real }}, y_{{real }}))] \\
&+\mathbb{E}_{z^{+} \sim Z^{+}}[\log (1-D_{q}(G_{\text {image }}(E(z^{+})), G_{\operatorname{mask}}(E(z^{+}))))]
\end{aligned}
\end{equation}
\noindent \textbf{Summarized Advantages: 1) Lightweight.} Our $G_{mask}(\cdot)$ is extremely lightweight yet powerful, which consists of a OAFF and classification head with total 90K parameters and 3.6MB model size. \textbf{2) Fewer labels.} We only need 522 images to train the $G_{mask}(\cdot)$ because our $G_{mask}(\cdot)$ is lightweight with only 90K parameters.

\subsection{Quality-aware Discriminator}

Our $D_{q}(\cdot)$ can select high-quality synthesized image-mask pairs from noisy synthetic data pool, providing high-quality synthetic data to train the saliency network. We noticed that the synthetic data fails occasionally for non-rigid objects (e.g., dogs) due to their various poses, resulting in low-quality image-mask pairs. To alleviate this issue, we proposed a quality discriminator $D_{q}(\cdot)$ adopting the lightweight MobileNetV3 \cite{howard2019searching} as backbone, which aims to select high-quality synthesized image-mask pair. During training, we feed two pairs to the quality discriminator $D_{q}(\cdot)$, i.e., $(x_{real},y_{real})$ and $(x_{syn},y_{syn})$. Accordingly, the adversarial training loss for the $D_{q}(\cdot)$ can be formulated as:
\begin{equation}
\begin{aligned}
\mathcal{L}_{D_{q}}&=\operatorname{argmax} D_{q} \mathbb{E}_{(x_{{real}}, y_{ {real}}) \sim \mathcal{D}_{m}}[\log(D_{q}(x_{ {real}}, y_{ {real}}))] \\
&+\mathbb{E}_{z^{+} \sim Z^{+}}[\log (1-D_{q}(G_{\text {image}}(E(z^{+})), G_{{mask}}(E(z^{+}))))]
\end{aligned}
\end{equation}
Note that our $D_{q}(\cdot)$ is different from the typical discriminator, where the discriminator is designed for discriminating real or fake images, while our $D_{q}(\cdot)$ performs image-mask quality control.

\noindent \textbf{Summarized Advantages: 1) High-quality image-mask pairs.} Our SODGAN can generate any desired number of high-quality image-mask pairs, which forms our synthetic dataset. The generated image-mask pairs can then be used to train any off-the-shelf SOD architecture just like real datasets are. \textbf{2) Strong generalization capabilities.} Unlike previous works \cite{richter2016playing,ros2016synthia,wang2019learning,kar2019meta}, which usually arises significant domain gap between the synthetic (from computer games) and real-world domains, the presented SODGAN can generate realistic images (see Fig. \ref{fig:gan_pic}) and show strong generalization capabilities on the real test datasets (see Table \ref{tab:main_results}).

\section{Results and Analysis}
\subsection{Classification Head Architecture}
\label{cls_head}
In this section, we provide the detailed implementation regarding two aspects: convolutional neural networks and MLP. 

\noindent \textbf{CNN Architecture.} We first use a linear embedding layer to reduce the input dimension from $C$ to 128, followed by 3 convolutional layers with kernel size of 3. The corresponding dimensions of the output channels are 128, 32, and 2 (the number of classes). All the layers are followed by a leaky ReLU activation function except for the last output layer. We call this standard version CNN-S. We also introduce CNN-M and CNN-L, where M/L denotes medium/large model size, and the architecture hyper-parameters of these model variants can be seen in the first 2 rows of Table \ref{tab:net_detail}.

\noindent \textbf{MLP Architecture.} We build our base model, called MLP-S, which consists of 3 fully-connected layers with 128, 32, and 2 hidden nodes, respectively. All layers except the output layer are followed by the BatchNorm layer and ReLU activation function. Similar to CNN-S, we also introduce its variants version MLP-M and MLP-L, and their hyper-parameters can be seen in the last 2 rows of Table \ref{tab:net_detail}.

\begin{table}[!h]
\center{
\normalsize{
\linespread{2}
\renewcommand\arraystretch{1}
\resizebox{0.75\linewidth}{!}{
\begin{tabular}{l||c|c}
\Xhline{1pt}
 & Layers & Channels\\
\hline
\hline
CNN-M & 5 & $\{128, 64, 64, 32, 2\}$\\
CNN-L & 7 & $\{128, 64, 64, 64, 64, 32, 2\}$\\
\hline
MLP-M & 4 & $\{128, 64, 32, 2\}$\\
MLP-L & 5 & $\{128, 64, 64, 32, 2\}$\\
\Xhline{1pt}
\end{tabular}}}}

\caption{Architecture details for the adopted CNN/MLP.}
\label{tab:net_detail}
\end{table} \vspace{-1em}

\subsection{Synthetic Data VS. Real DUTS-TR}

\begin{figure*}[t]
\centering
\includegraphics[width=0.95\textwidth]{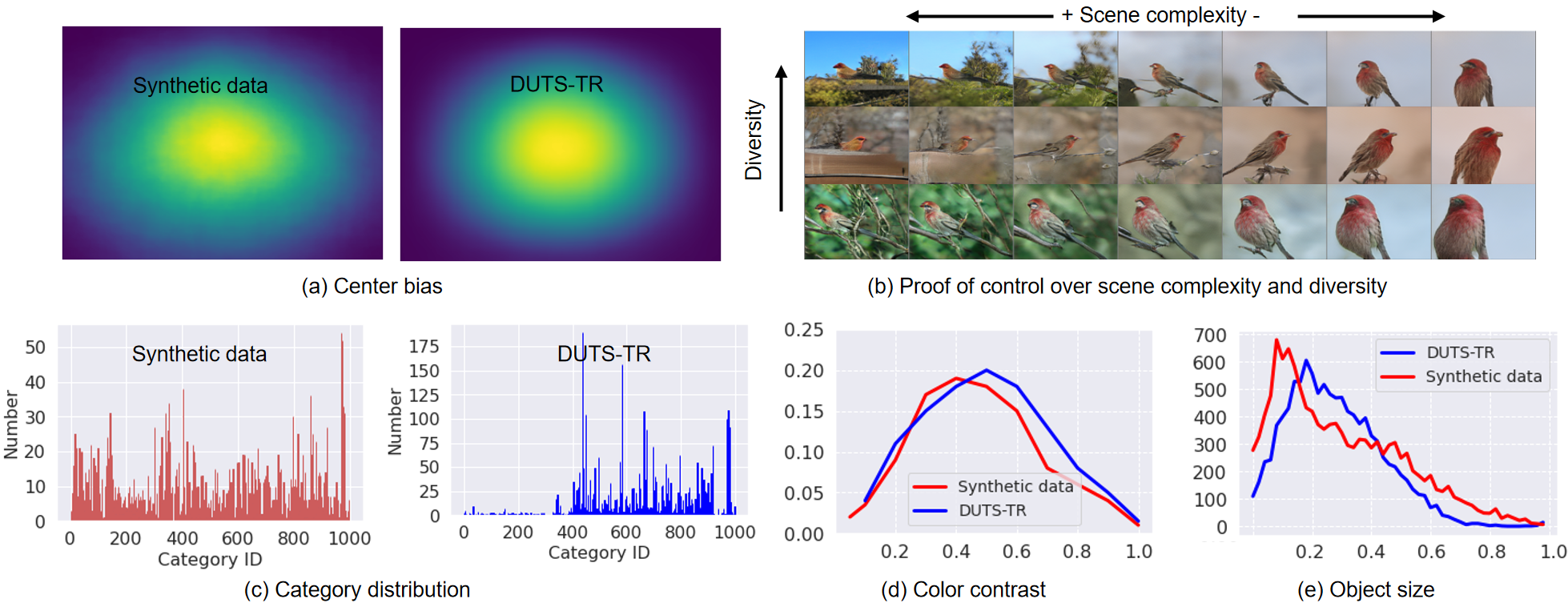}
\vspace{-1em}
\caption{Extensive analysis of synthetic data and the real DUTS-TR dataset show that the synthetic data have many advantages over the DUTS-TR in terms of center bias, category distribution, color contrast, and salient object size. Besides, the synthetic data can control the samples complexity and diversity.}
\label{fig:distribution_analysis}
\end{figure*}

As shown in Fig. \ref{fig:distribution_analysis}, we provide analyses of our synthesized datasets compared to the real DUTS-TR datasets in terms of center bias, category distribution, color contrast, and salient object size.

\begin{table}[!t]
\center{
\Large{
\renewcommand\arraystretch{0.9}
\resizebox{0.9\linewidth}{!}{
\begin{tabular}{l||ccc|ccc}
\Xhline{0.8pt}
\multirow{2}{*}{}
& \multicolumn{3}{c|}{DUTS-TE}
& \multicolumn{3}{c}{ECSSD}
\\
\cline{2-7}
   &max$F_\beta\uparrow$  & S-m$\uparrow$ & MAE$\downarrow$    &max$F_\beta\uparrow$  & S-m$\uparrow$ &MAE$\downarrow$     \\
\hline\hline
VAEs & .8375 & .8331 & .0644 & .9241 & .8945 & .0466 \\
$DEN(\cdot)$ & \textbf{.8557} & \textbf{.8507} & \textbf{.0530} & \textbf{.9377} & \textbf{.9129} & \textbf{.0389} \\
\hline\hline

w/o OAFF & .8333 & .8320& .0622 & .9187 & .8945 &.0476 \\
w/ $GA(\cdot)$ & .8437 & .8413 & .0594 & 9285 & .9024 & .0429 \\
w/ OAFF &  \textbf{.8557} & \textbf{.8507} & \textbf{.0530} & \textbf{.9377} & \textbf{.9129} & \textbf{.0389} \\
\hline
 CNN-S  &.8416 & .8400 & .0562 &.9229 & .8995 & .0464    \\
 CNN-M & \textbf{.8424} & \textbf{.8478}& \textbf{.0479} & \textbf{.9274} & \textbf{.9025} & \textbf{.0432}    \\
CNN-L  & .8208 &  .8098 &.0787 & .9206 & .8905 & .0529    \\
\hline
  MLP-S  & \textbf{.8557} &  \textbf{.8507} & \textbf{.0530} & \textbf{.9377} & \textbf{.9129} & \textbf{.0389}    \\
MLP-M   &.8382 & .8327   & .0625  &  .9349 & .9118  & .0399     \\
MLP-L & .8510  & .8476   & .0551  & .9375   & .9114 & .0394     \\
\hline\hline
w/o $D_q$  &.8406  & .8370   & .0603  & .9306   & .9038 &.0437     \\
 w/ $D_q$ & \textbf{.8557} &  \textbf{.8507} & \textbf{.0530} & \textbf{.9377} & \textbf{.9129} &  \textbf{.0389}    \\
\Xhline{0.8pt}
 \end{tabular}  }} 
\caption{Comparisons of different network structures for mask generator on DUTS-TE and ECSSD datasets.}
\label{tab:ablation_study}
}\vspace{-1em}
\end{table}

\noindent \textbf{Center bias.} We visualize the salient object locations for the synthetic data and the DUTS-TR datasets in Fig. \ref{fig:distribution_analysis}.a. Most objects are biased towards the image center for both datasets. Compared to the DUTS-TR, the synthetic data show lower center distributions. \textbf{Category distribution.} We use the SOTA classification model CoAtNet \cite{dai2021coatnet} to classify the filtered synthetic data and the DUTS-TR, which can be divided into 764 and 522 categories, respectively. As shown in Fig. \ref{fig:distribution_analysis}.c, our synthetic data contains more object categories than the DUTS-TR. \textbf{Color contrast \& Object size.} Since the DUTS-TR was designed for SOD tasks, the DUTS-TR’s images containing at least one salient object are higher color contrast than randomly generated synthetic data (see Fig. \ref{fig:distribution_analysis}.d). Besides, we also statistics the object size of the DUTS-TR and our synthetic data in Fig. \ref{fig:distribution_analysis}.e. As we can see, the synthetic data also contains smaller objects than the DUTS-TR. Additionally, BigGAN introduced the “truncation coefficient” $\lambda$, allowing explicit, fine-grained control of the trade-off between sample variety and complexity (see Fig. \ref{fig:distribution_analysis}.b).

\subsection{Ablation Study of Our Innovations}

\noindent \textbf{Eeffects of the proposed $DEN(\cdot)$.} To demonstrate the effects of our $DEN(\cdot)$, we compared the proposed $DEN(\cdot)$ with commonly used VAEs. As shown in Table \ref{tab:ablation_study}, the proposed $DEN(\cdot)$ improved by $1.8\%$ compared to the VAEs in terms of S-measure, which shows the effectiveness of the proposed diffusion model.

 \begin{figure}[!h]
  \centering
    \centering
    \includegraphics[width=0.495\linewidth]{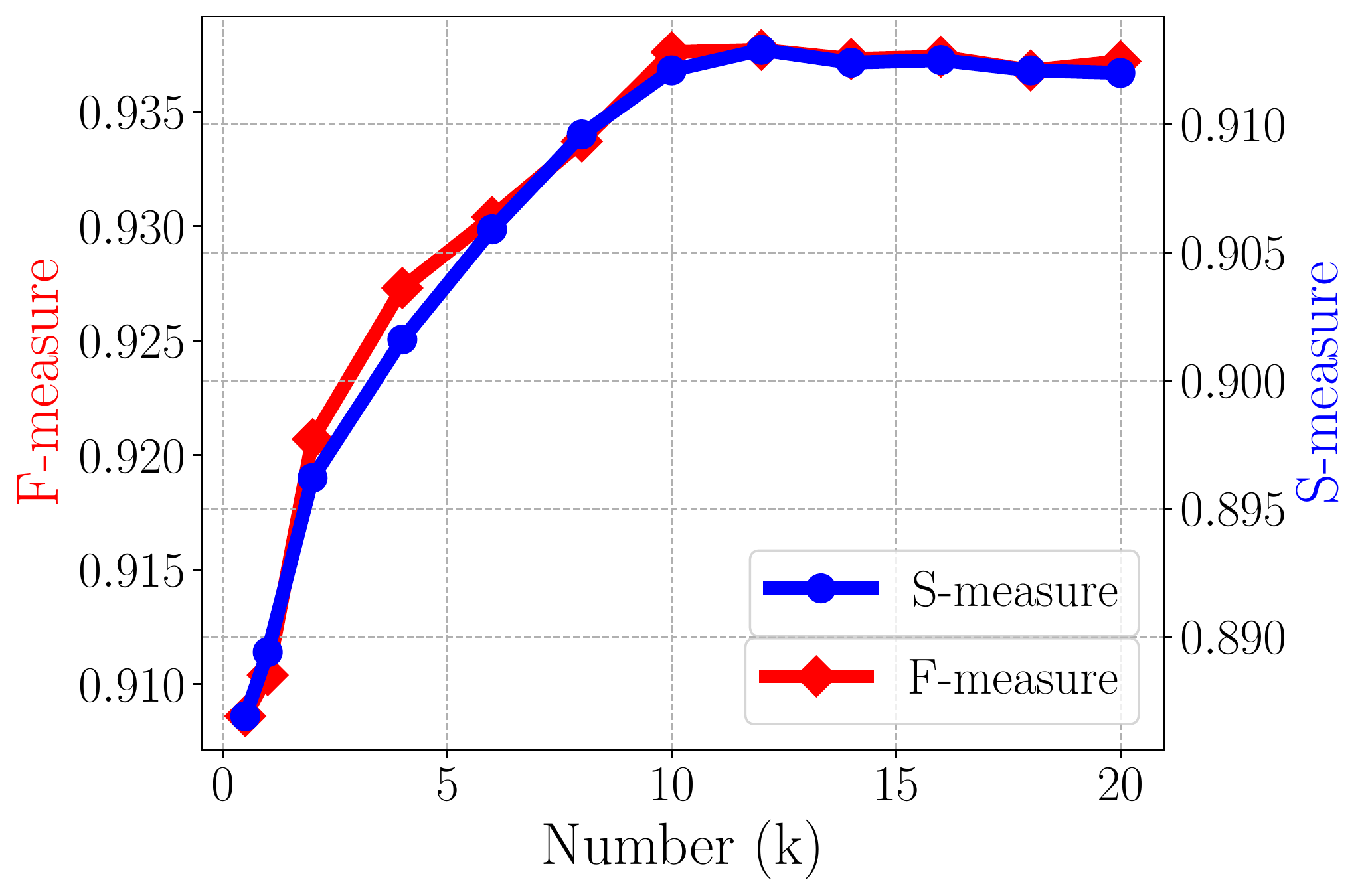}
    \includegraphics[width=0.495\linewidth]{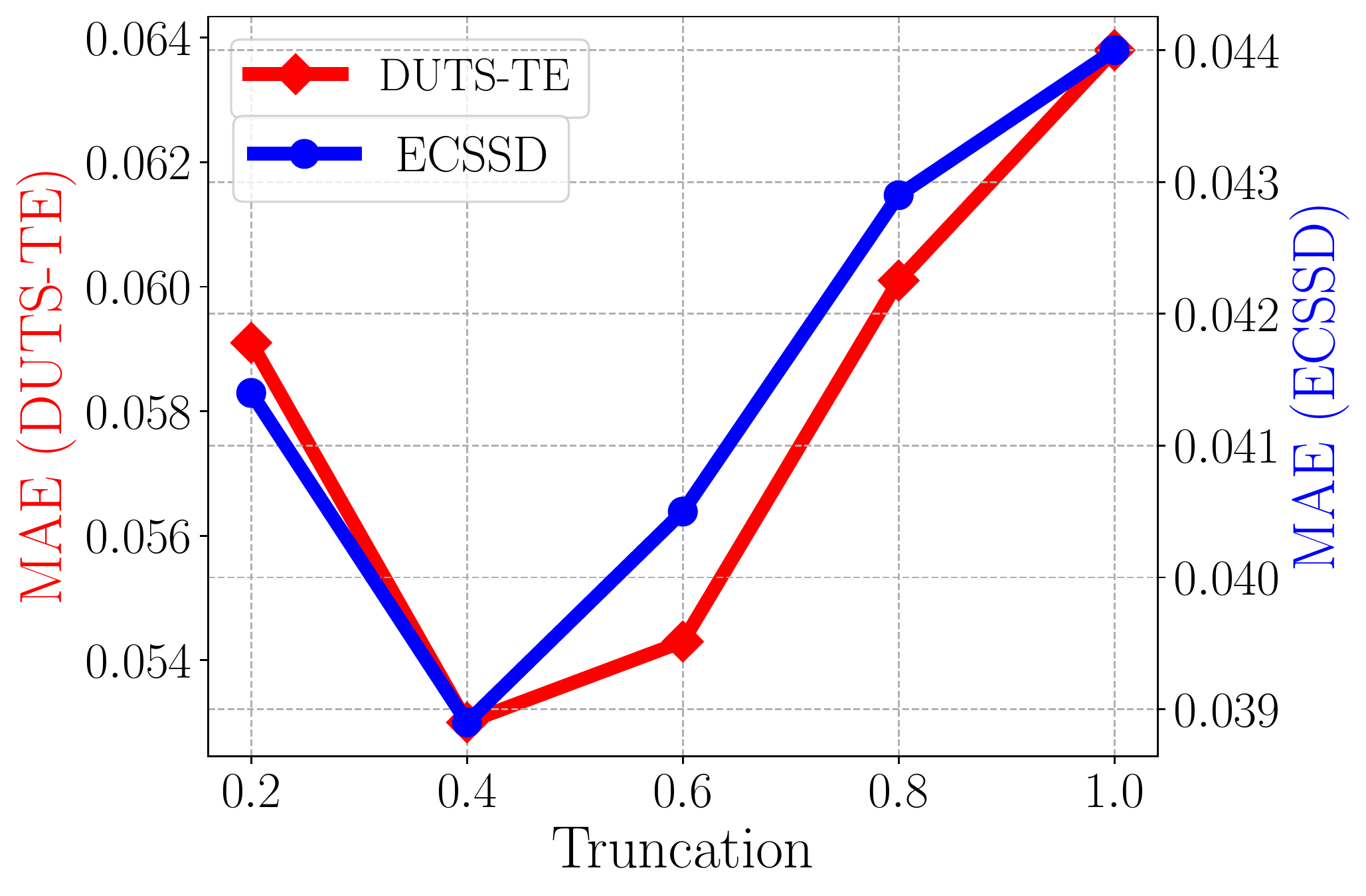}\vspace{-1em}
  \caption{Left: Results on varying amount of synthesized data. Right: The effects of different truncation $\lambda$.}
  \label{fig:lambda_truncation} \vspace{-1em}
\end{figure}

\noindent \textbf{Eeffects of the proposed OAFF.} In Table \ref{tab:ablation_study}, we evaluate 3 settings of OAFF: \textbf{1)} $G_{mask}(\cdot)$ without using the OAFF; \textbf{2)} $G_{mask}(\cdot)$ only using the global attention $GA(\cdot)$; \textbf{3)} the $G_{mask}(\cdot)$ with the OAFF. As we can see, the $G_{mask}(\cdot)$ with the $GA(\cdot)$ achieves better performance than the plain version, and the performance can be further improved by using OAFF, demonstrating the contribution of the OAFF to the final results.

\noindent \textbf{The choice of classification head architecture.} We evaluate 2 architectures on the proposed classification head network, i.e., CNN and MLP, with small (S), medium (M), and large (L) networks described in Sec. \ref{cls_head}. As shown in Table \ref{tab:ablation_study}, we notice that the MLP-S outperforms all the three CNN networks. Besides, we also notice that smaller networks obtain better performance due to the limited training data. Therefore, we take the MLP-S with channel dimension \{128, 32, 2\} as our classification head.

\noindent \textbf{Eeffects of the proposed $D_q(\cdot)$.} To illustrate the effectiveness of the proposed $D_q(\cdot)$, we implement 2 different settings, i.e., our SODGAN with/without using the $D_q(\cdot)$. As shown in Table \ref{tab:ablation_study}, the performance can be improved by $1.5\%$ in terms of F-measure on the DUTS-TE dataset by using the $D_q(\cdot)$, verifying the contribution of our $D_q(\cdot)$ to the final results.

\begin{table*}[t]
\centering
\Large{
\renewcommand\arraystretch{0.75}
\resizebox{0.95\textwidth}{!}{
\begin{tabular}{c|r||cccccccc||ccccccc}
\Xhline{1pt} \rule{0pt}{11pt}            &        & \multicolumn{8}{c||}{\textbf{Fully-Supervised Models}}                                                      & \multicolumn{7}{c}{\textbf{Semi/Weakly-Supervised Models}}                              \\ \cline{3-17}
           & Metric & DGRL    & PAGR    & BAS     & CPD     & MINet   & F3Net           & SAMN           & PFSN           & MWS     & ENDS           & WS$^3$A    & SCWS           & FCS & MFNet     & \multirow{2}{*}{\textbf{Ours}}           \\
           &        & \cite{DGRL18} & \cite{PAGRN} & \cite{BASNet19} & \cite{CPD19} &\cite{pang2020multi} & \cite{wei2020f3net}        & \cite{liu2021samnet}        &  \cite{ma2021PFSNet}        & \cite{zeng2019multi} & \cite{zhang2020learning}        & \cite{zhang2020weakly} & \cite{yu2021structure}        & \cite{zhang2021few} & \cite{piao2021mfnet}                \\ \hline\hline
\multirow{5}{*}{\rotatebox{90}{{DUTS-OM}.}}
           & maxF$\uparrow $  & .7742   & .7709   & .8053   & .7966   & .8098   & \textbf{\textcolor{blue}{.8133}}           & .8026         & \textbf{\textcolor{red}{.8233}}  & .7176   & .7581          & .7532   & \textbf{\textcolor{blue}{.7827}}          & .7170   & .7062 & \textbf{\textcolor{red}{.7930}} \\

           & S-m$\uparrow $   & .8059   & .7751   & .8362   & .8248   & .8329   & \textbf{\textcolor{blue}{.8385}}           & .8299         & \textbf{\textcolor{red}{.8425}}  & .7559   & .7832          & .7848   & \textbf{\textcolor{blue}{.8019}}          & .7448   & .7418 &  \textbf{\textcolor{red}{.8022}} \\

           & MAE$\downarrow$    & .0618   & .0709   & .0565   & .0560   & .0555   & \textbf{\textcolor{red}{.0526}}  & .0652         & \textbf{\textcolor{blue}{.0545}}           & .1086   & .0759          & .0684   & \textbf{\textcolor{red}{.0602}}          & \textbf{\textcolor{blue}{.0656}}   & .0867 & {.0768} \\

           & AUC$\uparrow $   & .8821   & .8983   & .9262   & .9378   & .9396   & .9413           & \textbf{\textcolor{red}{.9573}}& \textbf{\textcolor{blue}{.9496}}           & .9413   & \textbf{\textcolor{blue}{.9506}} & .9182   & .8822          & .8381  & .9090 &  \textbf{\textcolor{red}{.9565}}          \\

           & avgF$\uparrow$   & .7656   & .7354   & .7875   & .7770   & .7907   & \textbf{\textcolor{blue}{.7957}}           & .7655         & \textbf{\textcolor{red}{.8069}}  & .6777   & .7246          & .7386   & \textbf{\textcolor{blue}{.7602}} & .7073   & .6816 &  \textbf{\textcolor{red}{.7689}}          \\ \Xhline{0.6pt}

\multirow{5}{*}{\rotatebox{90}{{DUTS-TE}}}
           & maxF$\uparrow$   & .8287   & .8545   & .8591   & .8654   & .8835   & \textbf{\textcolor{blue}{.8905}}           & .8360         & \textbf{\textcolor{red}{.8949}}  & .7686   & .8173          & .7889   & \textbf{\textcolor{blue}{.8448}}          & .8296  &.7707 &  \textbf{\textcolor{red}{.8557}} \\

           & S-m$\uparrow$    & .8410   & .8369   & .8649   & .8684   & .8834   & \textbf{\textcolor{blue}{.8881}}           & .8479         & \textbf{\textcolor{red}{.8916}}  & .7573   & .8190          & .8021   & \textbf{\textcolor{blue}{.8391}}          & .8206   & .7728&  \textbf{\textcolor{red}{.8507}} \\

           & MAE$\downarrow$    & .0500   & .0562   & .0480   & .0438   & .0375   & \textbf{\textcolor{red}{.0358}}  & .0582         & \textbf{\textcolor{blue}{.0359}}           & .0920   & .0657          & .0628   & \textbf{\textcolor{blue}{.0493}}          & \textbf{\textcolor{red}{.0459}}  & .0772 &  {.0530} \\

           & AUC$\uparrow$      & .9137   & .9540   & .9451   & .9627   & .9714   & \textbf{\textcolor{blue}{.9726}}           & .9708         & \textbf{\textcolor{red}{.9739}}  & .9539   & \textbf{\textcolor{blue}{.9645}}          & .9312   & .8967          & .9000   &.9343 &  \textbf{\textcolor{red}{.9720}} \\

           & avgF$\uparrow$     & .8209   & .8108   & .8261   & .8357   & .8566   & \textbf{\textcolor{blue}{.8647}}           & .7920         & \textbf{\textcolor{red}{.8714}}  & .7311   & .7743          & .7715   & \textbf{\textcolor{blue}{.8326}}          & .8085   & .7415 &  \textbf{\textcolor{red}{.8374}} \\ \Xhline{0.6pt}

\multirow{5}{*}{\rotatebox{90}{{ECSSD}}}
           & maxF$\uparrow$     & .9224   & .9268   & .9424   & .9392   & \textbf{\textcolor{blue}{.9475}}   & .9453           & .9279         & \textbf{\textcolor{red}{.9523}}  & .8778   & .9002          & .8880   & \textbf{\textcolor{blue}{.9145}}          & .9108   & .8796 & \textbf{\textcolor{red}{.9377}} \\

           & S-m$\uparrow$      & .9028   & .8892   & .9162   & .9181   & \textbf{\textcolor{blue}{.9249}}   & .9242           & .9071         & \textbf{\textcolor{red}{.9298}}  & .8275   & .8707          & .8655   & \textbf{\textcolor{blue}{.8818}}          & .8787   & .8345 & \textbf{\textcolor{red}{.9129}} \\

           & MAE$\downarrow$      & .0407   & .0609   & .0370   & .0371   & .0334   & \textbf{\textcolor{blue}{.0333}}           & .0501         & \textbf{\textcolor{red}{.0309}}  & .0963   & .0676          & .0590   & .0489          & \textbf{\textcolor{blue}{.0471}}   & .0843 & \textbf{\textcolor{red}{.0389}} \\

           & AUC$\uparrow$      & .9505   & .9685   & .9666   & .9812   & .9845   & .9846           & \textbf{\textcolor{blue}{.9857}}         & \textbf{\textcolor{red}{.9860}}  & .9771   & \textbf{\textcolor{blue}{.9776}}          & .9531   & .9268          & .9478  &  .9497 &  \textbf{\textcolor{red}{.9868}} \\

           & avgF$\uparrow$     & .9122   & .8944   & .8970   & .9216   & \textbf{\textcolor{blue}{.9295}}   & .9272           & .8985         & \textbf{\textcolor{red}{.9346}}  & .8430   & .8730          & .8733   & \textbf{\textcolor{blue}{.9003}}          & .8951   & .8490 & \textbf{\textcolor{red}{.9137}} \\ \Xhline{0.6pt}

\multirow{5}{*}{\rotatebox{90}{{HKU-IS}}}
           & maxF$\uparrow$     & .9105   & .9176   & .9285   & .9251   & .9351   & \textbf{\textcolor{blue}{.9368}}           & .9147         & \textbf{\textcolor{red}{.9428}}  & .8560   & .9041          & .8805   & \textbf{\textcolor{blue}{.9085}}          & .8992   &  .8766 & \textbf{\textcolor{red}{.9320}} \\

           & S-m$\uparrow$      & .8945   & .8873   & .9090   & .9055   & .9190   & \textbf{\textcolor{blue}{.9173}}           & .8983         & \textbf{\textcolor{red}{.9244}}  & .8182   & \textbf{\textcolor{blue}{.8838}}          & .8649   & .8820          & .8718   & .8465& \textbf{\textcolor{red}{.9092}} \\

           & MAE$\downarrow$      & .0356   & .0475   & .0322   & .0342   & .0285   & \textbf{\textcolor{blue}{.0280}}           & .0449         & \textbf{\textcolor{red}{.0259}}  & .0843   & .0461          & .0470   & \textbf{\textcolor{blue}{.0375}}          & .0389   & .0585&  \textbf{\textcolor{red}{.0324}} \\

           & AUC$\uparrow$      & .9475   & .9704   & .9650   & .9765   & .9833   & .9817           & \textbf{\textcolor{red}{.9852}}         & \textbf{\textcolor{blue}{.9834}}  & \textbf{\textcolor{blue}{.9774}}   & .9826          & .9564   & .9282          & .9401   & .9671 & \textbf{\textcolor{red}{.9861}} \\

           & avgF$\uparrow$     & .8968   & .8904   & .9046   & .9004   & .9172   & \textbf{\textcolor{blue}{.9177}}           & .8856         & \textbf{\textcolor{red}{.9256}}  & .8291   & .8801          & .8677   & \textbf{\textcolor{blue}{.8945}}          & .8836   & .8535& \textbf{\textcolor{red}{.9054}} \\ \Xhline{0.6pt}

\multirow{5}{*}{\rotatebox{90}{{PASCAL-S}}}
           & maxF$\uparrow$     & .8808   & .8691   & .8757   & .8841   & .8894   & \textbf{\textcolor{blue}{.8948}}           & .8568         & \textbf{\textcolor{red}{.8986}}  & .8140   & .8706          & .8374   & .8660          & \textbf{\textcolor{blue}{.8742}}   & .8202 & \textbf{\textcolor{red}{.8924}} \\

           & S-m$\uparrow$      & .8278   & .7925   & .8194   & .8277   & .8333   & \textbf{\textcolor{blue}{.8404}}           & .8027         & \textbf{\textcolor{red}{.8431}}  & .7532   & .8025          & .7805   & .7936          & \textbf{\textcolor{blue}{.8102}}   & .7489 & \textbf{\textcolor{red}{.8422}} \\

          & MAE$\downarrow$      & .0823   & .1149   & .0924   & .0890   & .0828   & \textbf{\textcolor{blue}{.0799}}           & .1130         & \textbf{\textcolor{red}{.0790}}  & .1509   & .1144          & .1106   & .1000          & \textbf{\textcolor{blue}{.0849}}   & .1379 & \textbf{\textcolor{red}{.0743}} \\

           & AUC$\uparrow$      & .8988   & .9162   & .9113   & .9316   & .9339   & \textbf{\textcolor{blue}{.9428}}           & .9348         & \textbf{\textcolor{red}{.9438}}  & .9494   & \textbf{\textcolor{red}{.9588}} & .9062   & .8643          & .9071   & .9032 & \textbf{\textcolor{blue}{.9497}}          \\

           & avgF$\uparrow$     & .8528   & .8148   & .8100   & .8439   & .8512   & \textbf{\textcolor{blue}{.8580}}           & .8054         & \textbf{\textcolor{red}{.8614}}  & .7566   & .8222          & .8054   & .8321          & \textbf{\textcolor{blue}{.8387}}   & .7739&  \textbf{\textcolor{red}{.8542}} \\ \Xhline{1pt}
\end{tabular}
}}
\caption{Extensive experiments demonstrate that our approach achieves a new SOTA performance in terms of semi/weakly supervised methods, and even outperforms several fully-supervised SOTA methods. The top 2 results are highlighted in \textcolor{red}{red} and \textcolor{blue}{blue} respectively. The “DUTS-OM.” denotes DUT-OMRON dataset. The detailed training data setting can be found in Table \ref{tab:training_data}\vspace{-1em}}
\label{tab:main_results}
\end{table*}


\begin{figure*}[!t]

   \centering
   \resizebox{0.93\textwidth}{!}{

   \begin{subfigure}{.2\textwidth}
     \centering
     \includegraphics[width=\textwidth]{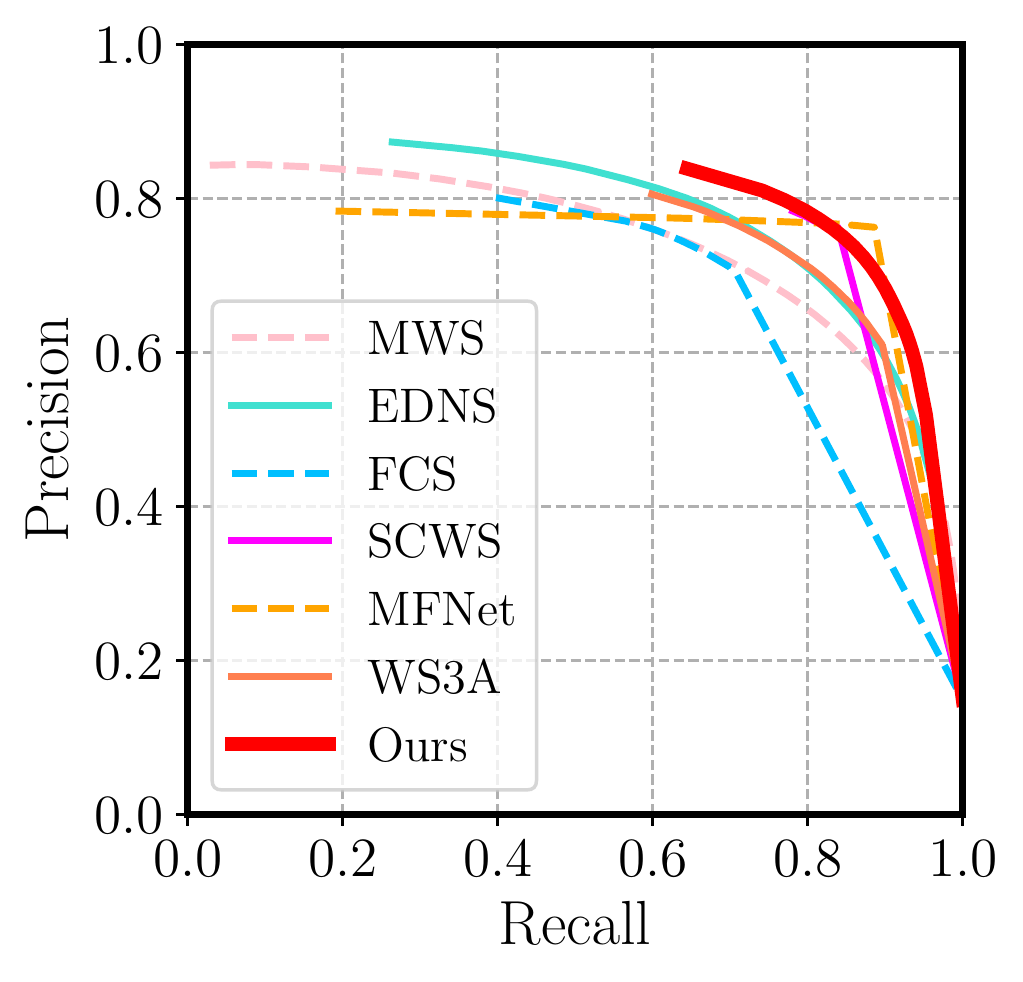}
   \end{subfigure}\hspace*{-0.50em}
   \begin{subfigure}{.2\textwidth}
     \centering
     \includegraphics[width=\textwidth]{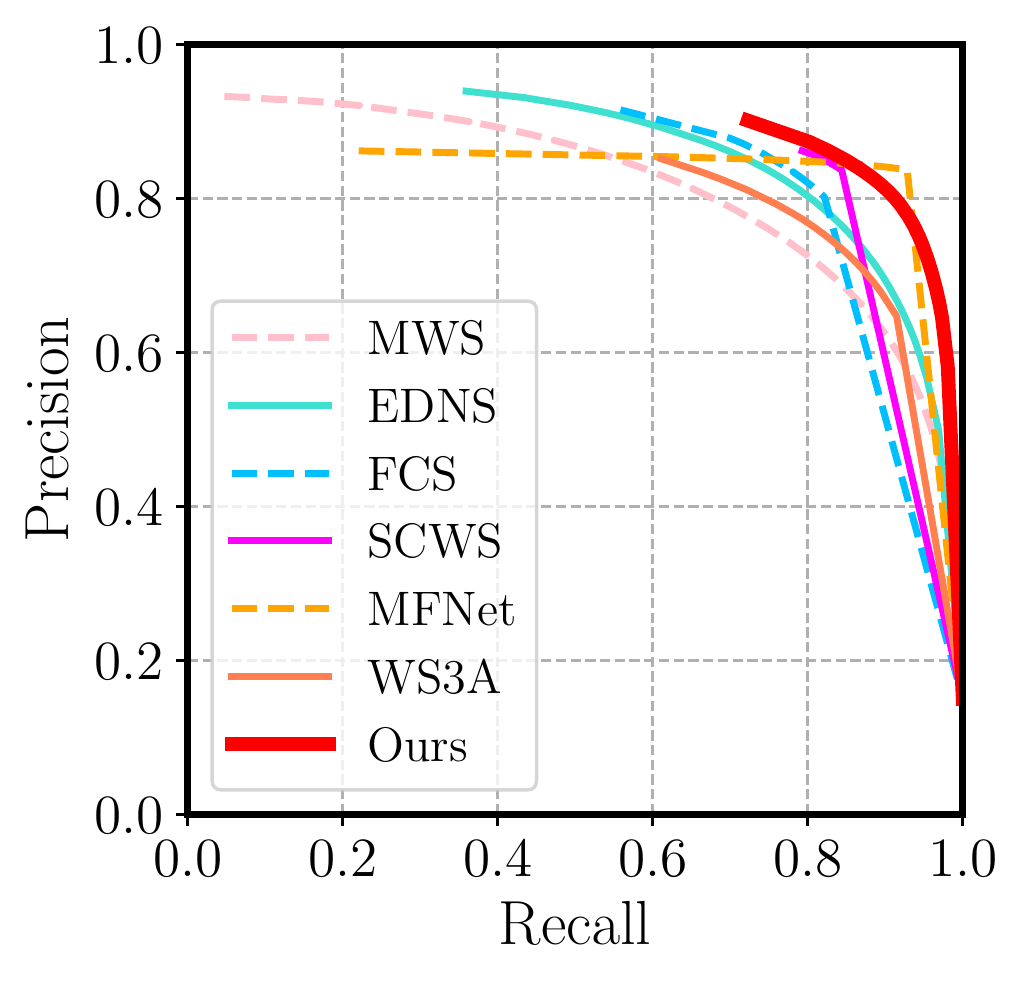}
   \end{subfigure}\hspace*{-0.50em}
    \begin{subfigure}{.2\textwidth}
     \centering
     \includegraphics[width=\textwidth]{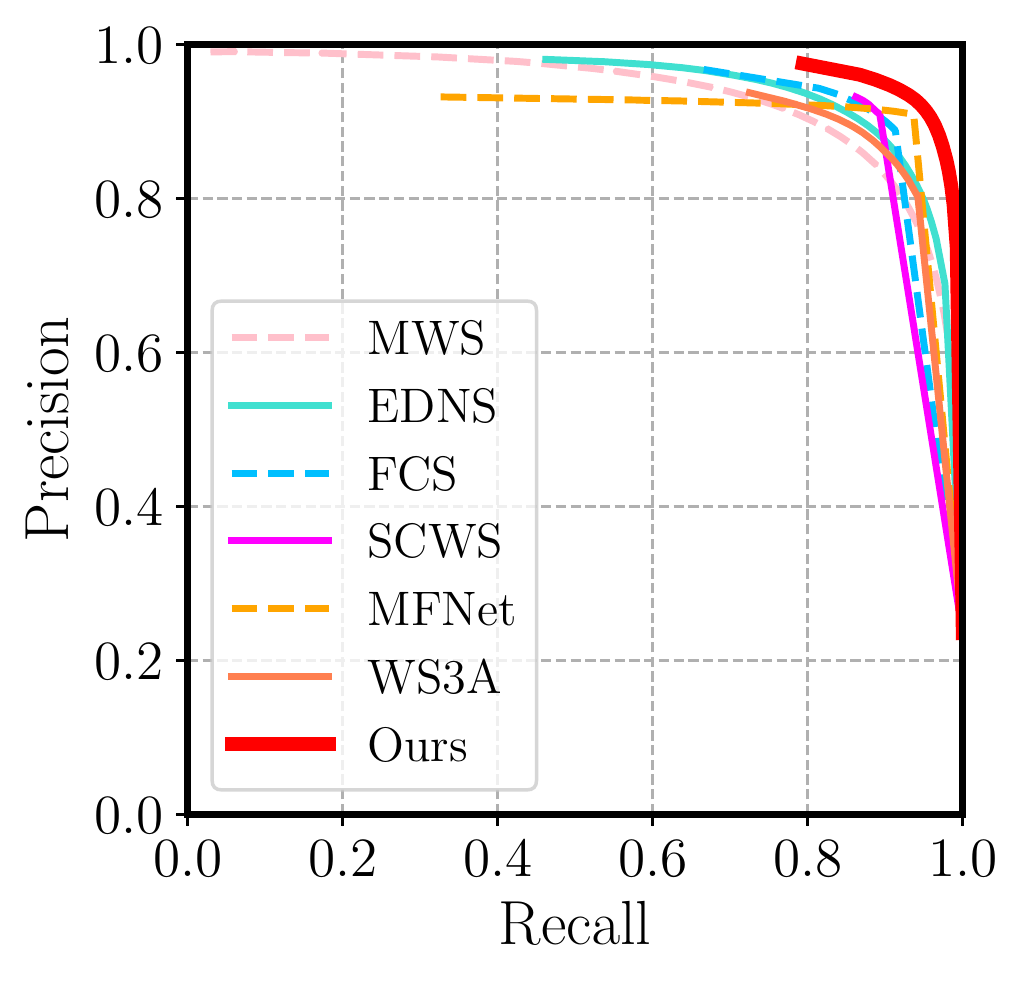}
   \end{subfigure}\hspace*{-0.50em}
    \begin{subfigure}{.2\textwidth}
     \centering
     \includegraphics[width=\textwidth]{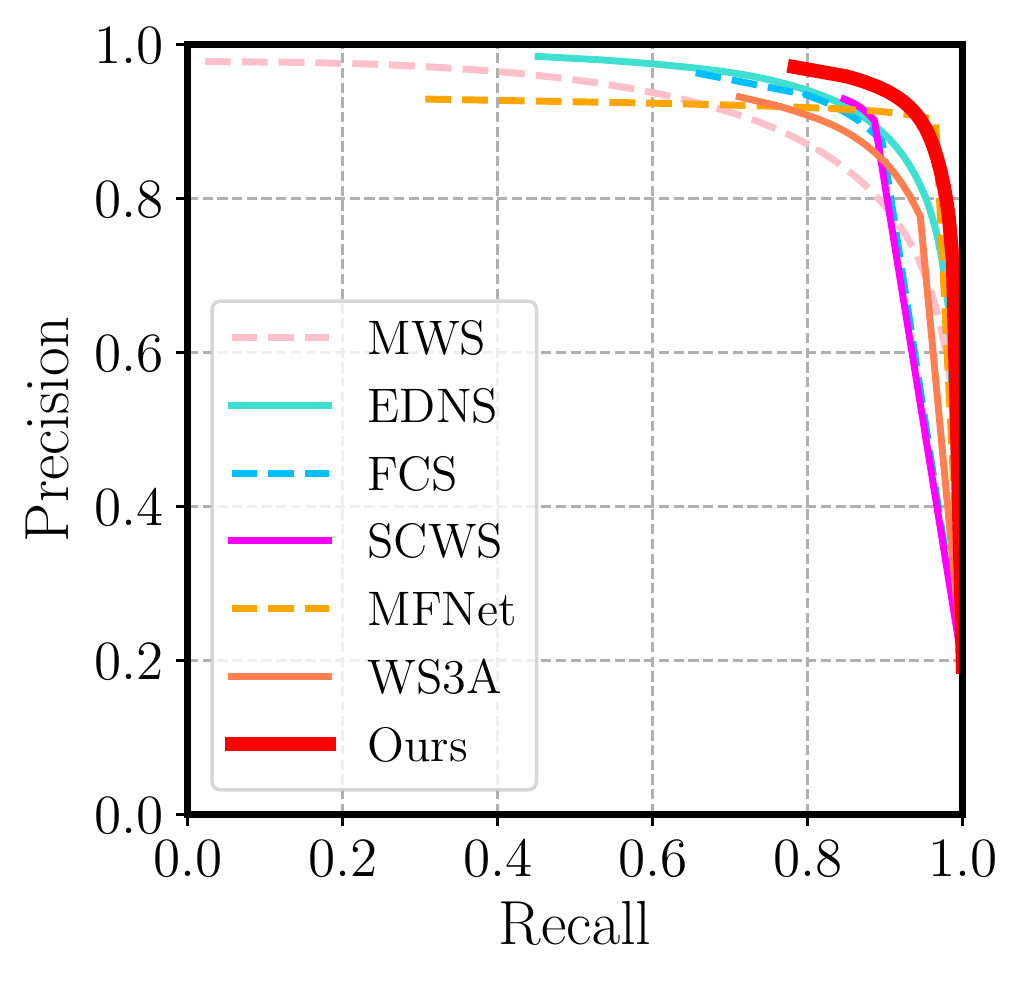}
   \end{subfigure}
    \begin{subfigure}{.2\textwidth}
     \centering
     \includegraphics[width=\textwidth]{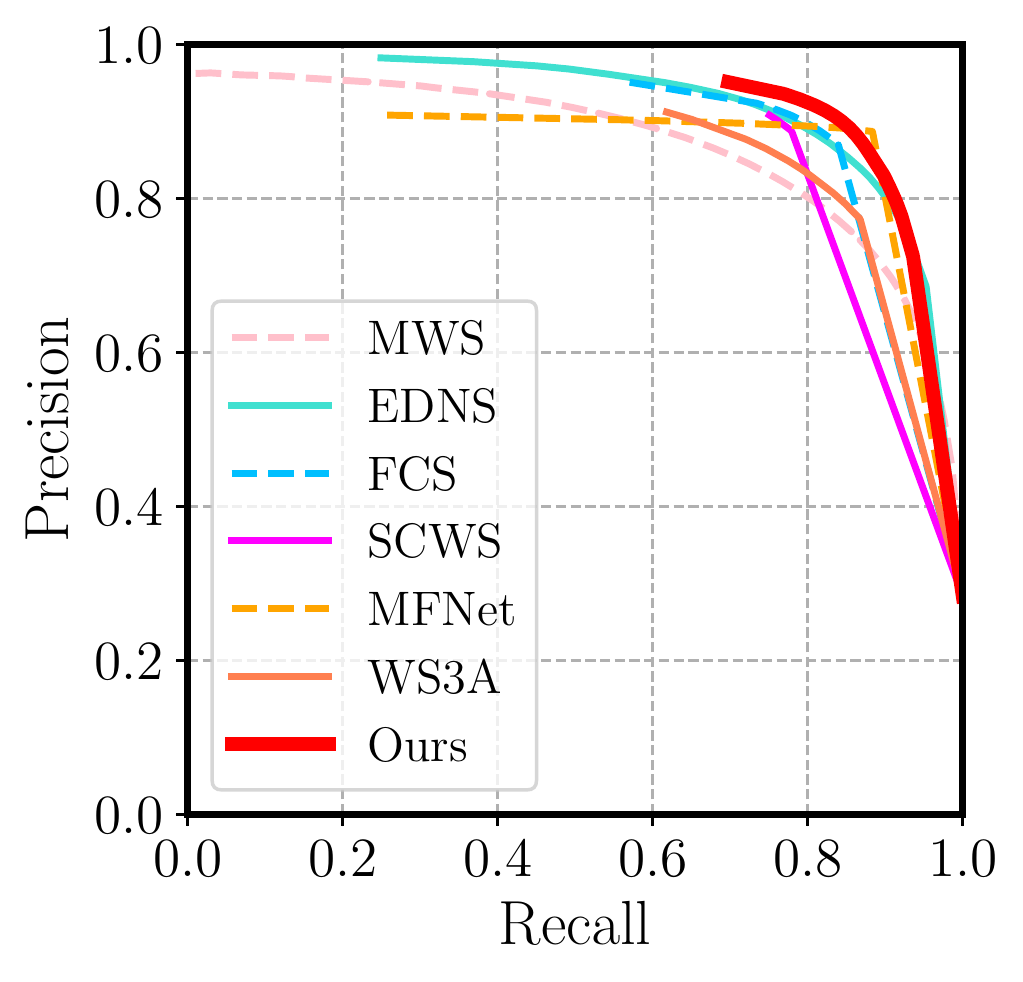}
   \end{subfigure}
   }\vspace{-0.4em}

   \resizebox{0.95\textwidth}{!}{

   \begin{subfigure}{.2\textwidth}
     \centering
     \includegraphics[width=\textwidth]{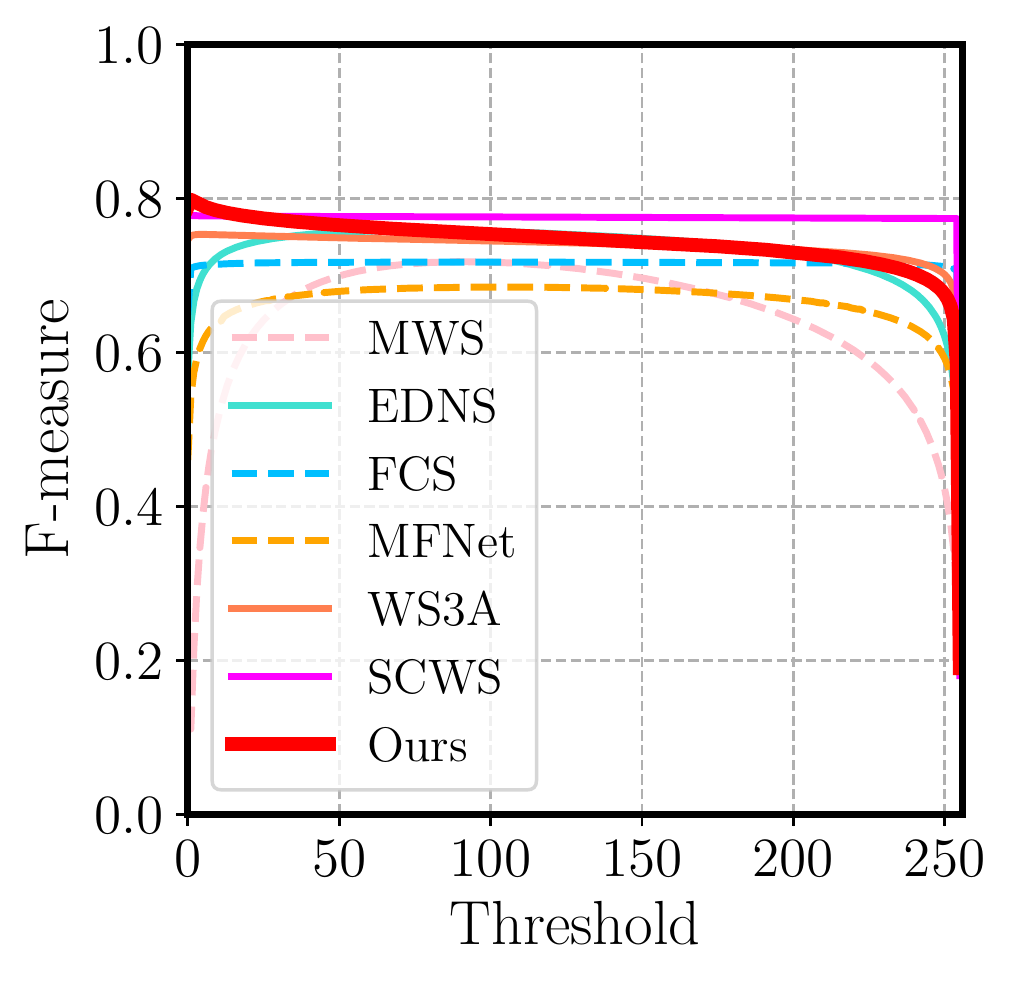}
     \caption{DUT-OMRON}
   \end{subfigure}\hspace*{-0.50em}
   \begin{subfigure}{.2\textwidth}
     \centering
     \includegraphics[width=\textwidth]{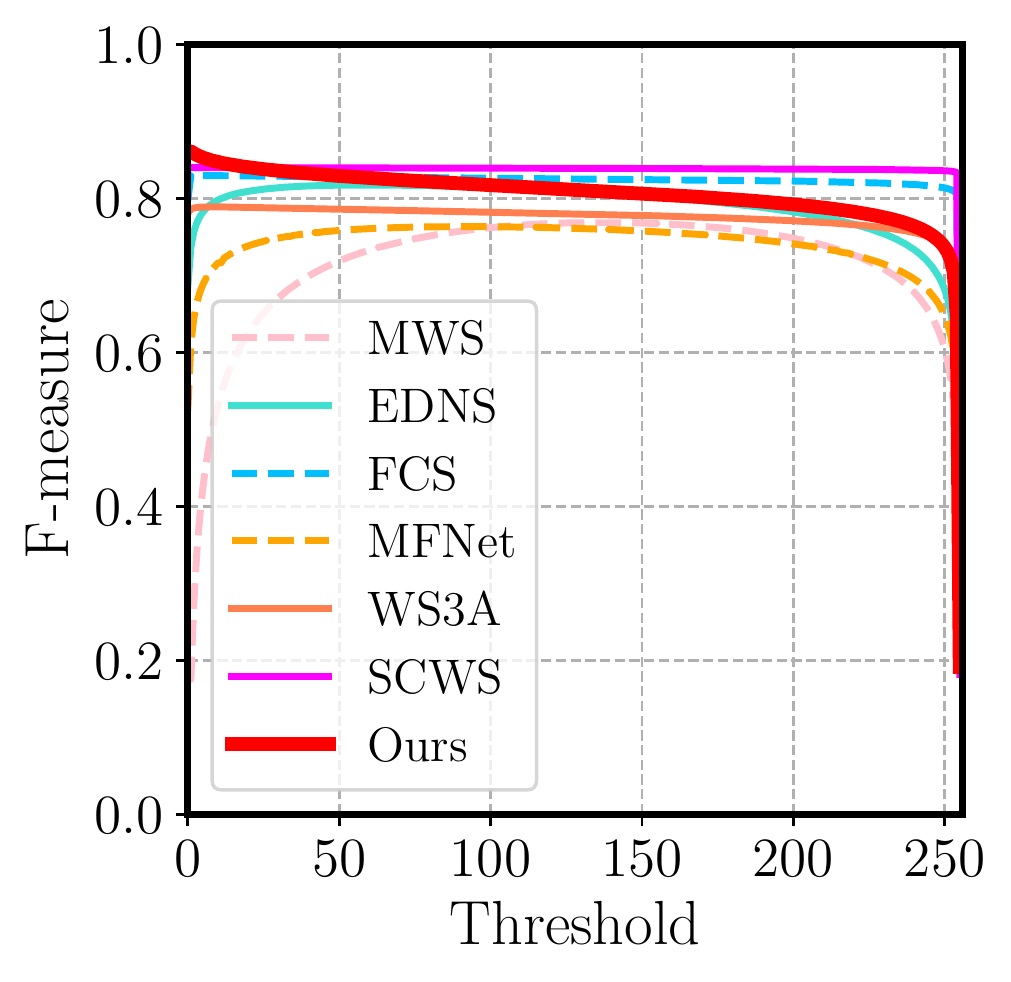}
     \caption{DUTS-TE}
   \end{subfigure}\hspace*{-0.50em}
    \begin{subfigure}{.2\textwidth}
     \centering
     \includegraphics[width=\textwidth]{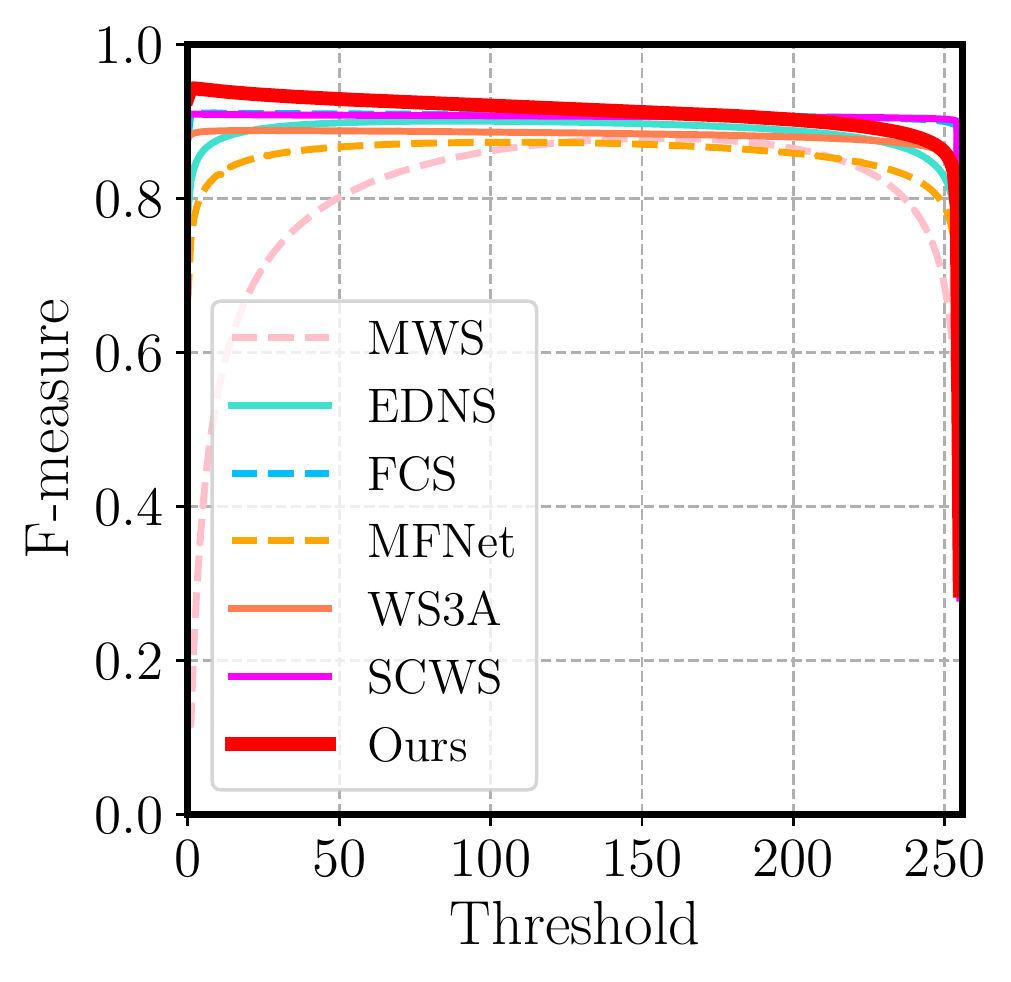}
     \caption{ECSSD}
   \end{subfigure}\hspace*{-0.50em}
    \begin{subfigure}{.2\textwidth}
     \centering
     \includegraphics[width=\textwidth]{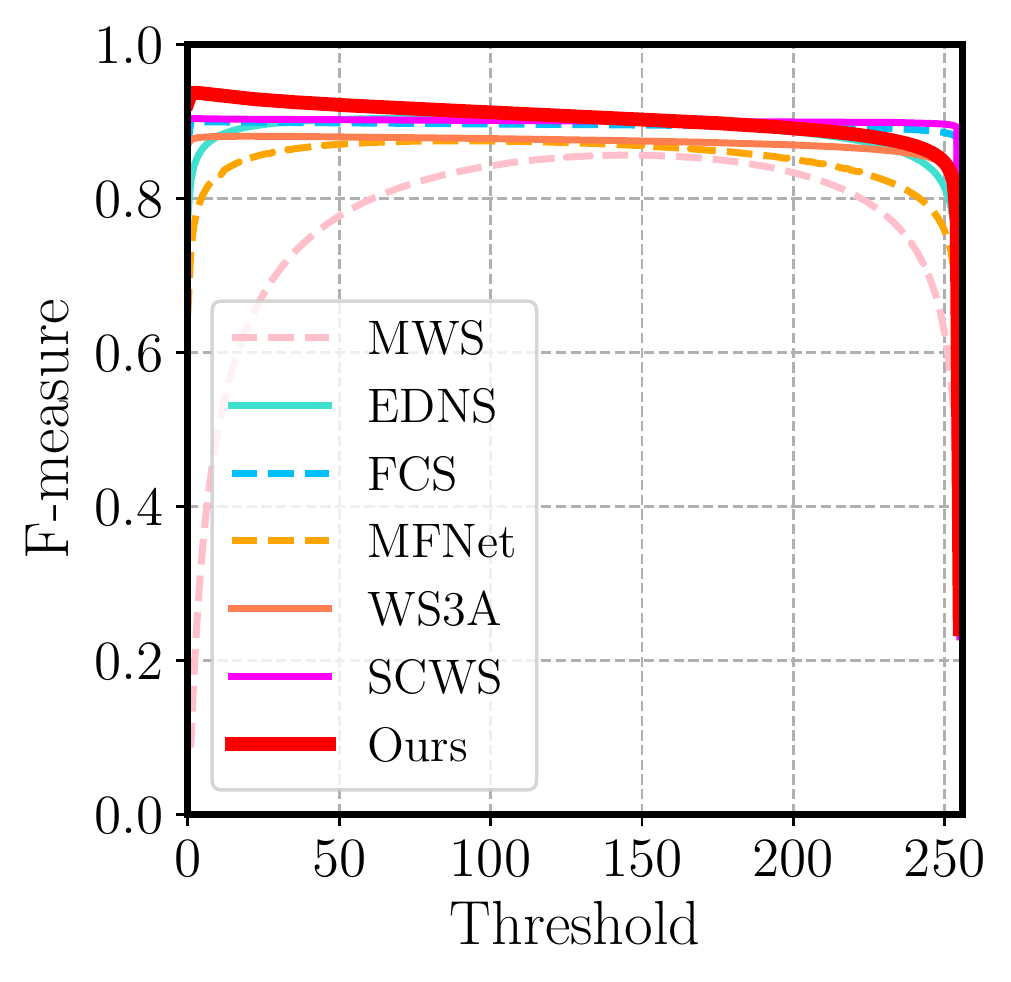}
     \caption{HKU-IS}
   \end{subfigure}
    \begin{subfigure}{.2\textwidth}
     \centering
     \includegraphics[width=\textwidth]{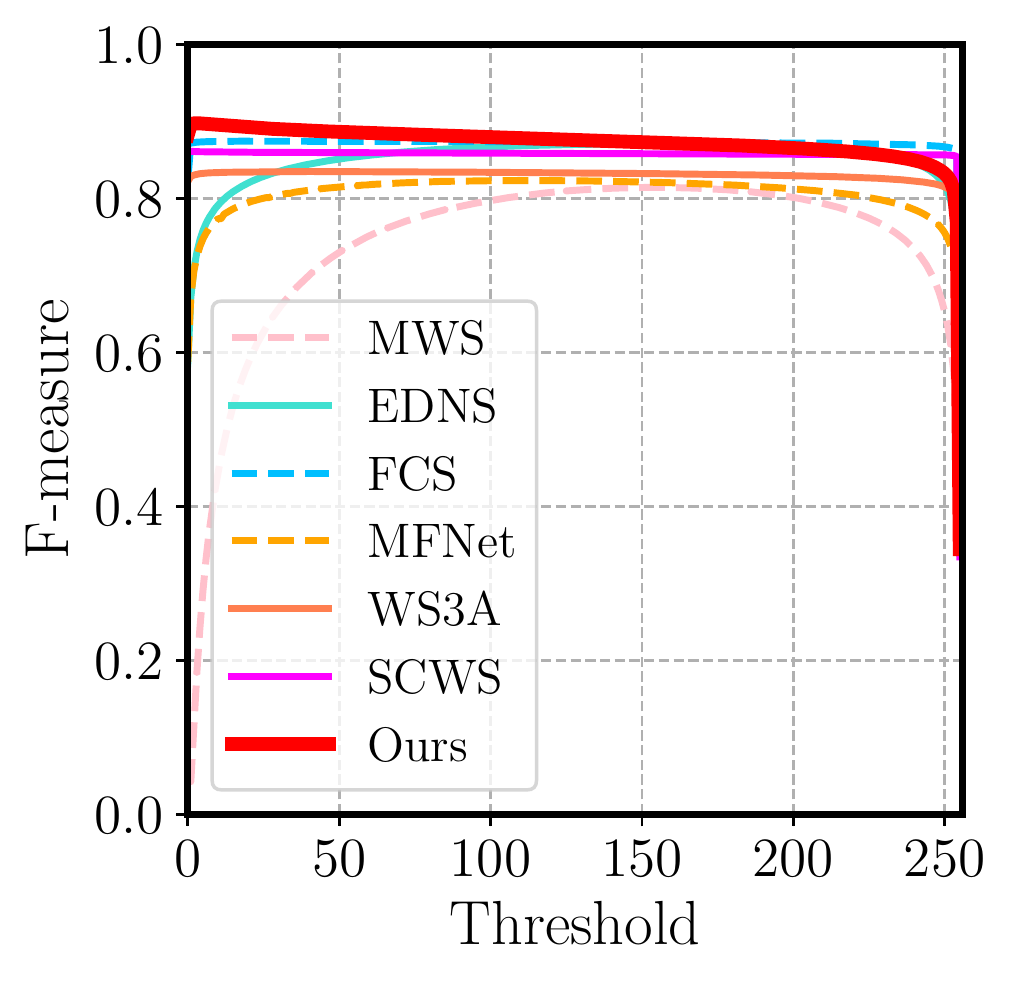}
     \caption{PASCAL-S}
   \end{subfigure}
   }\vspace{-1.2em}
   \caption{The performance on PR and F-measure curves also show the superiority of our method comparison with 6 SOTA weakly/semmi-supervised methods over 5 datasets.}
   \label{fig:pr_f_curves}
\end{figure*}


\noindent \textbf{Impacts of the amount of synthesized data.} We further explore the number of synthesized data how to influence the saliency performance. As shown in left of Fig. \ref{fig:lambda_truncation}, when the number of synthesized images is insufficient (< 12k), model performance can benefit substantially from the increased synthesized data. However, when the training set is large enough (> 12k), the application of more synthesized data does not necessarily lead to better performance. In this paper, unless otherwise specified, the reported SOD results were obtained by training on 12k synthetic image-mask pairs. Besides, to study the effects of $\lambda$, we vary the truncation coefficient $\lambda \in \{0.2, 0.4, 0.6, 0.8, 1\}$. The results are shown in the right of Fig. \ref{fig:lambda_truncation}. We observed that the saliency performance is inversely proportional to $\lambda$ when $\lambda > 0.4$, and the optimal setting is $\lambda =0.4$.

\subsection{Synthetic Data for SOD}

\noindent {\textbf{Setup.}} In this work, we do not focus on SOD network architecture design, so in our experiments, we adopt F3Net~\cite{wei2020f3net} as our saliency network by considering effectiveness and computational cost. Different from the previous works trained on the human wellannotated DUTS-TR \cite{wang2017learning} dataset (the detailed training data setting can be found in Table ~\ref{tab:training_data}), we train our model on the SODGAN’s generated images-mask pairs (12k).


\begin{figure*}[th]
\centering
\includegraphics[width=0.95\linewidth]{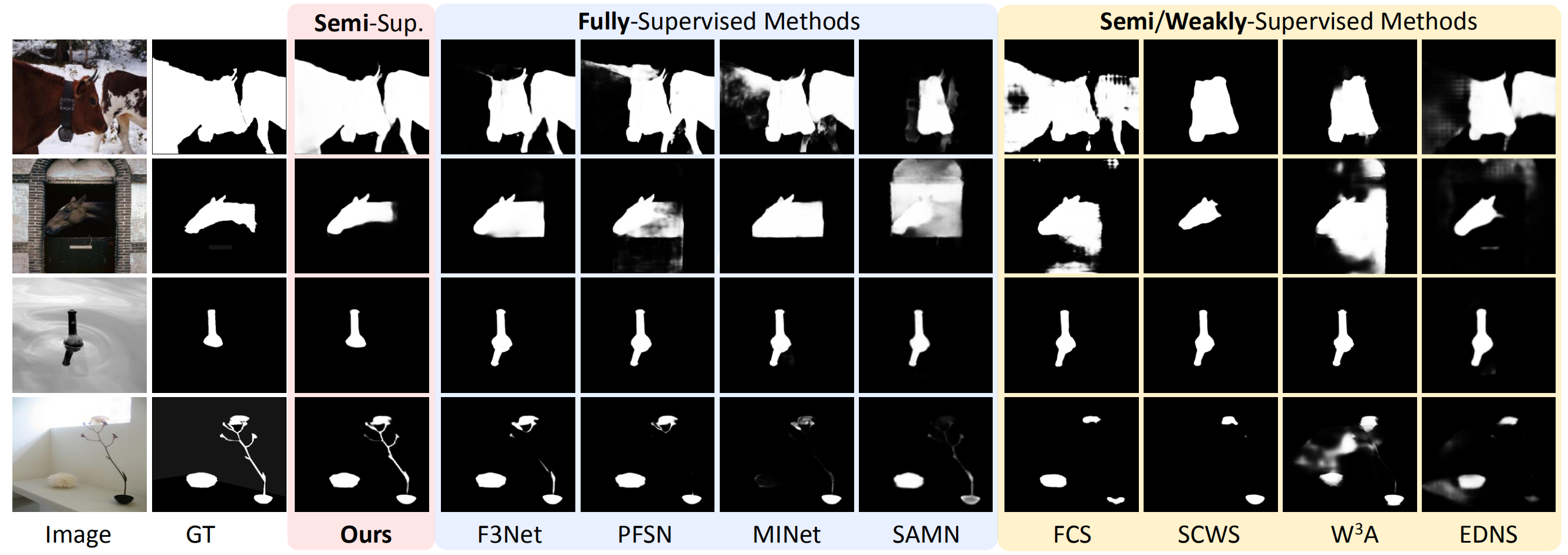}\vspace{-1em}
\caption{The visual comparison of the proposed model and existing SOTA methods also show that our model can generate more complete and accurate saliency maps than other semi/weakly-supervised, even outperforms fully-supervised SOD models.}
\label{fig:visual_pic}
\end{figure*}

\begin{table}[t]
\Large{
\renewcommand\arraystretch{0.7}
\centering
\resizebox{1\linewidth}{!}{
\begin{tabular}{c||c|c|c|r}
\Xhline{1pt}
Method      & Sup.  & Training dataset & Annotations      & Number    \\ \hline\hline
All-S       & F        & DUTS-TR          & Pixel-wise  & 10,553     \\ \hline
\multicolumn{1}{r||}{MWS~\cite{zeng2019multi}}         &W  & ImageNet+COCO   & Image-level      & 1.3M \\
\multicolumn{1}{r||} {MFNet~\cite{piao2021mfnet} } &    W    & ImageNet+DUTS-TR & Image-level   & 1.01M \\
\multicolumn{1}{r||}{EDNS~\cite{zhang2020learning}}       &W  & DUTS-TR          & Pseudo      & 10,553     \\
\multicolumn{1}{r||}{WS$^3$A~\cite{zhang2020weakly}}       & W & DUTS-TR          & Scribble    & 10,553     \\
\multicolumn{1}{r||}{SCWS~\cite{yu2021structure}} & W& DUTS-TR          & Scribble    & 10,553     \\ \hline
\multicolumn{1}{r||}{FCS~\cite{zhang2021few}}    &S        & DUTS-TR          & Pixel-wise  & 1,000      \\
\textbf{Ours}    &S         & DUTS-TR  & Pixel-wise  & 522       \\ \Xhline{1pt}
\end{tabular}
}}

\caption{Statistics of popular SOD training dataset. ``ALL-S'' denotes all supervised models in Tabel \ref{tab:main_results}. ``Sup.'' stands for level of supervision. ``F, W and S'' denote the fully-, weakly- and semi- supervised learning respectively. \vspace{-0.4cm}}
\label{tab:training_data}
\end{table}

\noindent {\textbf{Datasets.}} We evaluate the performance of the proposed method on 5 commonly used benchmark datasets, including DUTS-TE~\cite{wang2017learning}, DUT-OMRON ~\cite{yang2013saliency},  ECSSD~\cite{ecssd}, HKU-IS~\cite{zhao2015saliency}, and PASCAL-S~\cite{li2014secrets}.
\noindent {\textbf{Evaluation metrics.}} We adopt several widely-used metrics to evaluate our method, including the \textbf{P}recision-\textbf{R}ecall (PR) curves, the F-measure curves, Mean Absolute Error (MAE), max and mean F-measure \cite{w-fmeasure}, S-measure~\cite{Smeasure} and \textbf{A}rea \textbf{U}nder \textbf{C}urve (AUC).

\noindent {\textbf{Competitors.}} We compare the proposed approach with 13 SOTA SOD models, including MWS \cite{zeng2019multi}, EDNS \cite{zhang2020learning}, WS$^3$A \cite{zhang2020weakly}, SCWS \cite{yu2021structure}, FCS \cite{zhang2021few}, MFNet~\cite{piao2021mfnet}, {DGRL} \cite{DGRL18}, {PAGR}  \cite{PAGRN}, {BAS} \cite{BASNet19}, {CPD} \cite{CPD19}, {MINet} \cite{pang2020multi}, F3Net \cite{wei2020f3net}, PFSN~\cite{ma2021PFSNet}, and SAMN~\cite{liu2021samnet}. For fair comparison, we evaluate these SOTA models by using the same metric code with the authors provided saliency maps.

\noindent\textbf{Quantitative comparison.} In Table \ref{tab:main_results}, we compare our results with SOTA saliency methods. As indicated in Table \ref{tab:main_results}, our method consistently achieves significant improvement compared with semi- and weakly- supervised methods in terms of 5 evaluation metrics. Concretely, our method improved by $1.13\%$, $1.09\%$, $2.32\%$, $2.35\%$, and  $1.82\% $ on average compared to the second-best method in max F-measure on 5 datasets. Moreover, our saliency model even outperforms fully-supervised saliency models, such as CPD \cite{CPD19}, BAS \cite{BASNet19} and SAMN \cite{liu2021samnet}, on ECSSD, HKU-IS and PASCAL-S datasets. 
Our approach trained on synthetic data achieves comparable or superior to the fully supervised F3Net (0.8422 vs. 0.8404 in terms of S-measure on the PASCAL-S) trained on more than 10k well-annotated image-label pairs. Besides, we also provide the PR and F-measure curves in Fig. \ref{fig:pr_f_curves}, which also demonstrate the effectiveness of the synthesized high-quality image-mask pairs for saliency detection.

\noindent\textbf{Qualitative comparison.} As demonstrated in Fig. \ref{fig:visual_pic}, our synthetic data supervised saliency model has better visual superiority than other SOTA models. Concretely, our model excels in dealing with various challenging scenarios, including cluttered backgrounds (the 1st row), low contrast objects (the 2nd row), inverted reflection in the water (the 3rd row), and small objects (the 4th row). \vspace{-0.4cm}
\subsection{Conclusion}
In this paper, we present a simple but powerful approach, namely SODGAN, to explore the potential of synthetic data for SOD. It opens up a new research paradigm for semi-supervised SOD, and shows that promising segmentation accuracy can be achieved by using controllable synthesized data. Our major novelty is to discover the interpretable direction that can disentangle the foreground object from the background in GANs feature space with only a few annotated images. Our work expands the application of the generative model to salient object detection tasks. We believe this is only the first step by utilizing synthetic data to train saliency deep networks. 


\begin{acks}
This research is supported in part by the National Natural Science Foundation of China (No. 62172437 and 62172246), the Open Project
Program of State Key Laboratory of Virtual Reality Technology
and Systems (VRLAB2021A05), the Youth Innovation and Technology
Support Plan of Colleges and Universities in Shandong Province
(2021KJ062), and the Science and Technology Innovation Committee of
Shenzhen Municipality (No. JCYJ20210324131800002 and RCBS20210609103820029).
\end{acks}

\bibliographystyle{ACM-Reference-Format}
\bibliography{egbib}

\

\end{document}